\documentclass{article}

\usepackage[main, final]{neurips_2026}
\usepackage{graphicx}

\usepackage[utf8]{inputenc} 
\usepackage[T1]{fontenc}    
\usepackage{hyperref}       
\usepackage{url}            
\usepackage{booktabs}       
\usepackage{amsfonts}       
\usepackage{nicefrac}       
\usepackage{microtype}      
\usepackage[table]{xcolor}  

\usepackage[ruled,vlined,linesnumbered]{algorithm2e}
\usepackage{amsmath}
\usepackage{amssymb}
\usepackage{bm}
\usepackage{threeparttable}
\usepackage{multirow}

\title{EntropyCache: Decoded Token Entropy Guided KV Caching for Diffusion Language Models}

\author{%
  Minsoo Cheong \\
  Seoul National University\\
  \texttt{icycle0409@snu.ac.kr} \\
  \And
  Donghyun Son \\
  Stanford University\\
  \texttt{dhson@stanford.edu} \\
  \And
  Woosang Lim \\
  Seoul National University\\
  \texttt{ftyg656512@snu.ac.kr} \\
  \And
  Sungjoo Yoo\thanks{Corresponding author.} \\
  Seoul National University\\   
  \texttt{sungjoo.yoo@gmail.com} \\
}

\begin{document}

\maketitle

\begin{abstract}
Diffusion-based large language models (dLLMs) rely on bidirectional attention, which prevents lossless KV caching and requires a full forward pass at every denoising step. Existing approximate KV caching methods reduce this cost by selectively updating cached states, but their decision overhead scales with context length or model depth. We propose \textsc{EntropyCache}, a training-free KV caching method that uses the maximum entropy of newly decoded token distributions as a low-cost signal for deciding \emph{when} to recompute. Our design is grounded in two empirical observations: (1)~decoded token entropy correlates with KV cache drift, providing a cheap proxy for cache staleness, and (2)~feature volatility of decoded tokens persists for multiple steps after unmasking, motivating recomputation of the $k$ most recently decoded tokens.
For $N$ newly decoded tokens and vocabulary size $V$, the skip-or-recompute decision requires $O(NV)$ computation per step, independent of context length and model scale; in practice, $N$ is small and the measured overhead is only $0.5\%$ of inference time. Experiments on LLaDA-8B-Instruct and Dream-7B-Instruct show that \textsc{EntropyCache} achieves $15.2\times$--$26.4\times$ speedup on standard benchmarks and $22.4\times$--$24.1\times$ on chain-of-thought benchmarks against vanilla baselines, with competitive accuracy.
\end{abstract}

\section{Introduction}

Diffusion-based large language models (dLLMs)~\citep{d3pm, mdlm, llada, dream} have emerged as a compelling alternative to autoregressive (AR) models~\citep{llama}, enabling parallel token generation through iterative denoising. Rather than producing tokens sequentially from left to right, dLLMs operate over the entire output sequence simultaneously, progressively replacing mask tokens with predicted values across multiple denoising steps. When combined with semi-autoregressive generation strategies such as blockwise decoding~\citep{fastdllm} or sliding window decoding~\citep{elasticcache}, dLLMs have shown promising potential to achieve competitive throughput compared to conventional AR generation.

These gains, however, are constrained by a fundamental architectural limitation: dLLMs employ non-causal (bidirectional) attention. In AR models, causal attention~\citep{transformer, decoder_only_causal_attention} guarantees that previously computed key-value (KV) pairs remain the same as new tokens are appended, making KV caching~\citep{kv_cache} both straightforward and lossless. In dLLMs, every position attends to every other position; consequently, unmasking even a single token alters the representations of \emph{all} positions. Therefore, vanilla dLLM inference executes a full forward pass over the entire sequence at every denoising step, yielding a per-step cost proportional to the context length.

To reduce this cost, prior studies~\citep{dkv-cache, dllm-cache, flashdlm, fastdllm} observe that the KV states of most tokens change only minimally between consecutive denoising steps, suggesting that selective reuse of cached states can closely approximate exact recomputation. Building on this insight, several approximate KV caching strategies have been proposed. Fast-dLLM~\citep{fastdllm} takes a static approach by freezing the KV pairs outside the current denoising block throughout its denoising steps, sacrificing cache accuracy in favor of computational efficiency.

On the other hand, some studies propose to dynamically decide where or when to compute. For example, $\text{d}^2\text{Cache}$~\citep{d2cache} uses attention rollout to selectively update high-influence token positions at each step, focusing on \emph{where} along the token axis to recompute. Elastic-Cache~\citep{elasticcache} monitors per-layer attention-weight drift and recomputes from a boundary layer onward, indirectly addressing \emph{when} cached states have become stale. However, its detection requires per-layer cosine similarity comparison and attention score recomputation at every step, with decision overhead that scales with model depth, context length, and hidden dimension.

We propose EntropyCache, which provides a cheaper signal for deciding \emph{when} to recompute: the maximum entropy of the newly decoded token distributions. We show that this single scalar correlates with KV cache drift, enabling a skip-or-recompute decision whose overhead is independent of context length and model scale. The design is motivated by two empirical observations:

\textbf{Observation 1: Unmasked token entropy predicts KV cache drift.} We show that the maximum entropy~\citep{entropy} of the token distributions obtained from the previous denoising step correlates with the magnitude of the resulting KV cache drift at the subsequent step. This single scalar therefore serves as a cheap yet effective proxy for cache staleness.

\textbf{Observation 2: Feature volatility persists for multiple steps after unmasking.} Tracking individual token trajectories reveals that some tokens remain unstable for several denoising steps after being unmasked, not just the single step assumed by prior dynamic methods~\citep{dkv-cache, d2cache}. This motivates recomputing the $k$ most recently unmasked tokens rather than only those from the previous step.

Concretely, EntropyCache works as follows. After each unmasking step, it evaluates the maximum entropy of the newly unmasked token distributions. If this entropy exceeds a threshold $\tau$, a full forward pass is triggered; otherwise, the method reuses cached KV pairs for most positions and recomputes KV pairs only for (a) the current mask tokens and (b) at most $k$ recently unmasked tokens since the previous full recomputation. We make the following contributions:
\begin{itemize}
\item We identify maximum unmasked-token entropy as a lightweight, empirically grounded proxy for KV cache staleness, replacing expensive per-layer rollout or hidden-state comparisons with an $O(NV)$ computation over the $N$ newly decoded distributions, independent of context length, model depth, and hidden dimension.
\item We provide empirical analysis showing that feature volatility persists for multiple denoising steps after unmasking, and characterize how recency-ranked local recomputation effectively captures this multi-step instability.
\item We integrate these findings into EntropyCache, a training-free caching method for dLLMs. Evaluated on LLaDA-8B-Instruct and Dream-7B-Instruct, EntropyCache achieves \textbf{15.2$\times$}--\textbf{26.4$\times$} speedup on standard benchmarks and \textbf{22.4$\times$}--\textbf{24.1$\times$} on chain-of-thought benchmarks, consistently outperforming prior caching methods in throughput while maintaining competitive accuracy.
\end{itemize}

\section{Related work}

\subsection{Diffusion-based large language models}
Unlike autoregressive (AR) models that generate tokens left-to-right with causal attention~\citep{decoder_only_causal_attention,llama}, dLLMs denoise the entire sequence simultaneously using bidirectional attention. LLaDA~\citep{llada} trains a masked diffusion model from scratch and achieves performance competitive with LLaMA3-8B. Dream~\citep{dream} instead initializes from a pretrained AR checkpoint and introduces context-adaptive noise rescheduling. In both models, unmasking even a single token changes representations at all positions, making standard lossless KV caching inapplicable.
Vanilla dLLM inference runs a full forward pass at every denoising step, so cost scales with the product of sequence length and step count. Semi-autoregressive strategies reduce this cost. Blockwise decoding~\citep{fastdllm} denoises fixed-size blocks one at a time, while sliding-window decoding~\citep{elasticcache} advances a window over the sequence and keeps the number of denoised masked tokens constant at each step, better preserving diffusion-level parallelism. Orthogonally, confidence-based parallel decoding~\citep{fastdllm} unmasks multiple tokens per step when predicted probabilities exceed a threshold. These strategies are complementary to KV caching methods.

\subsection{Approximate KV caching for dLLMs}
Prior approximate KV caching strategies for dLLMs fall into two categories: static methods that apply a fixed reuse policy across all denoising steps, and dynamic methods that determine where or when to refresh the cache based on intermediate states at each step.

\paragraph{Static caching strategy.}
Fast-dLLM~\citep{fastdllm} freezes the KV pairs outside the current denoising block throughout its denoising steps. 
This is simple and effective but does not dynamically detect cache staleness. The entire cache is refreshed only at block boundaries.

\paragraph{Dynamic caching strategy.}
$\text{d}^2\text{Cache}$~\citep{d2cache} uses attention rollout to identify high-influence tokens and selectively updates their KV states at each step. 
Elastic-Cache~\citep{elasticcache} monitors attention-weight drift at the most-attended token and, upon detecting staleness, recomputes from a boundary layer onward while reusing shallow-layer caches. 
Both improve upon static caching in accuracy--throughput tradeoff, but their decision overhead scales with context length or model depth, limiting practical speedup.

\section{Motivation}
\label{sec:motivation}

\subsection{Decoded token entropy predicts KV cache drift}
\label{sec:entropy_kv_drift}
\begin{figure}[ht!]  
    \centering
    \includegraphics[width=\textwidth] 
    {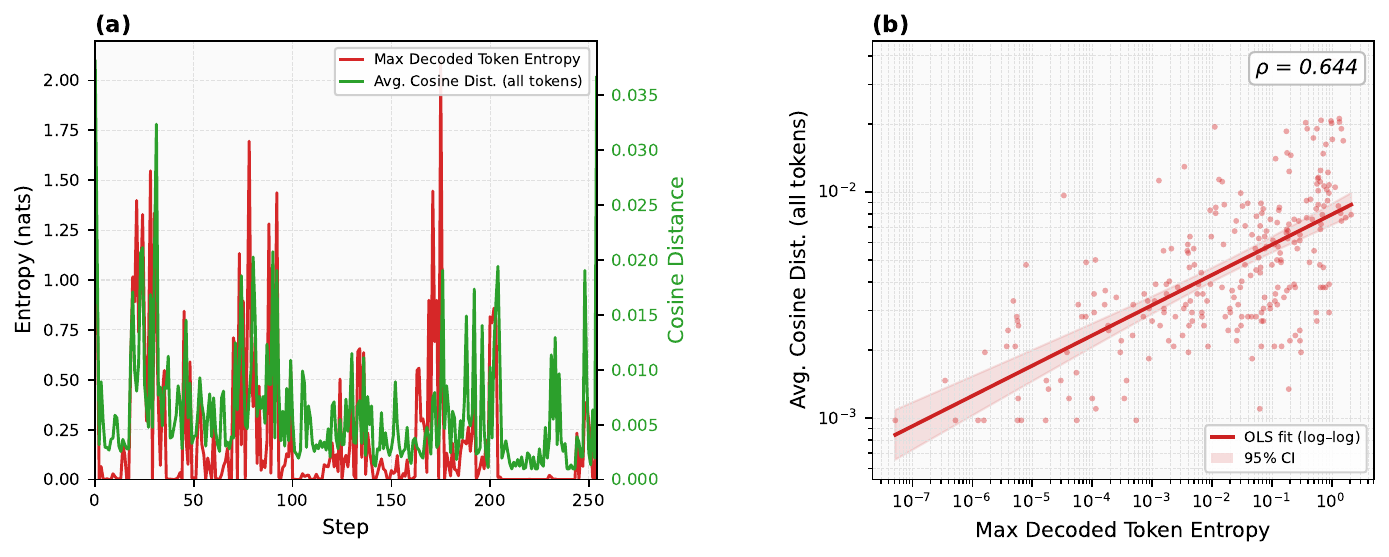}
    \caption{(a) Decoded-token entropy and value-vector cosine distance over denoising steps for a single GSM8K sample with LLaDA-8B-Instruct.
    (b) Maximum decoded-token entropy versus average value-vector cosine distance on log--log axes.}
    \label{fig:figure1}
\end{figure}
Entropy is a widely-used measure of predictive uncertainty in neural networks~\citep{entropy,dropout,xu2025gift}. 
We examine whether this signal can also serve as a predictor of KV cache drift in dLLMs, by measuring the relationship between prediction uncertainty at each denoising step and the resulting shift in the model's internal representations. Our central claim is that the \textbf{maximum entropy of the newly decoded token distributions} at each step serves as an effective, low-cost predictor of how much the KV cache will drift at the subsequent step, providing an effective proxy for selective recomputation.

We run LLaDA-8B-Instruct on a single GSM8K sample with 256 denoising steps, recording two quantities at every step: (i)~the maximum entropy among the newly decoded tokens and (ii)~the average cosine distance between value vectors at consecutive denoising steps ($t$ and $t{+}1$) across all token positions, which we take as our ground-truth measure of KV cache drift. We present this single-sample analysis here to build intuition; an extended study aggregating 64 samples per benchmark across four datasets, which confirms and strengthens these findings, is provided in Appendix~\ref{app:extended_entropy_analysis}.
\paragraph{Temporal co-occurrence of entropy spikes and cache drift.}
Figure~\ref{fig:figure1}(a) plots these two signals over the course of generation. The max decoded token entropy tracks the cosine distance spikes remarkably well: steps at which a high-entropy token is committed coincide with pronounced jumps in KV drift, whereas low-entropy decodings are followed by minimal KV drift. This suggests that cache staleness is driven specifically by the surprise introduced at the moment of token commitment.
\paragraph{Quantitative correlation.}
To move beyond visual inspection, we scatter-plot the max decoded token entropy against the average value-vector cosine distance for all 256 steps on log--log axes (Figure~\ref{fig:figure1}(b)). The two quantities exhibit a clear positive correlation in log--log space with a Spearman rank correlation of $\rho = 0.644$. The extended analysis shows that this trend holds consistently across tasks and samples (Appendix~\ref{app:extended_entropy_analysis}).

\paragraph{Intuition.}
We hypothesize that this relationship stems from how dLLMs propagate local changes in the input. When a mask token is resolved from a high-entropy (i.e., highly uncertain) distribution, the change in input sequence may inject a large amount of new information or surprise into the sequence. Conversely, when a token is decoded from a near-deterministic distribution, the model has already anticipated that outcome in its representations, and the resulting perturbation is small.
\subsection{Feature volatility persists beyond the decoding step}
\label{sec:recent_token_volatility}
\begin{figure}[ht!]  
    \centering
    \includegraphics[width=\textwidth]{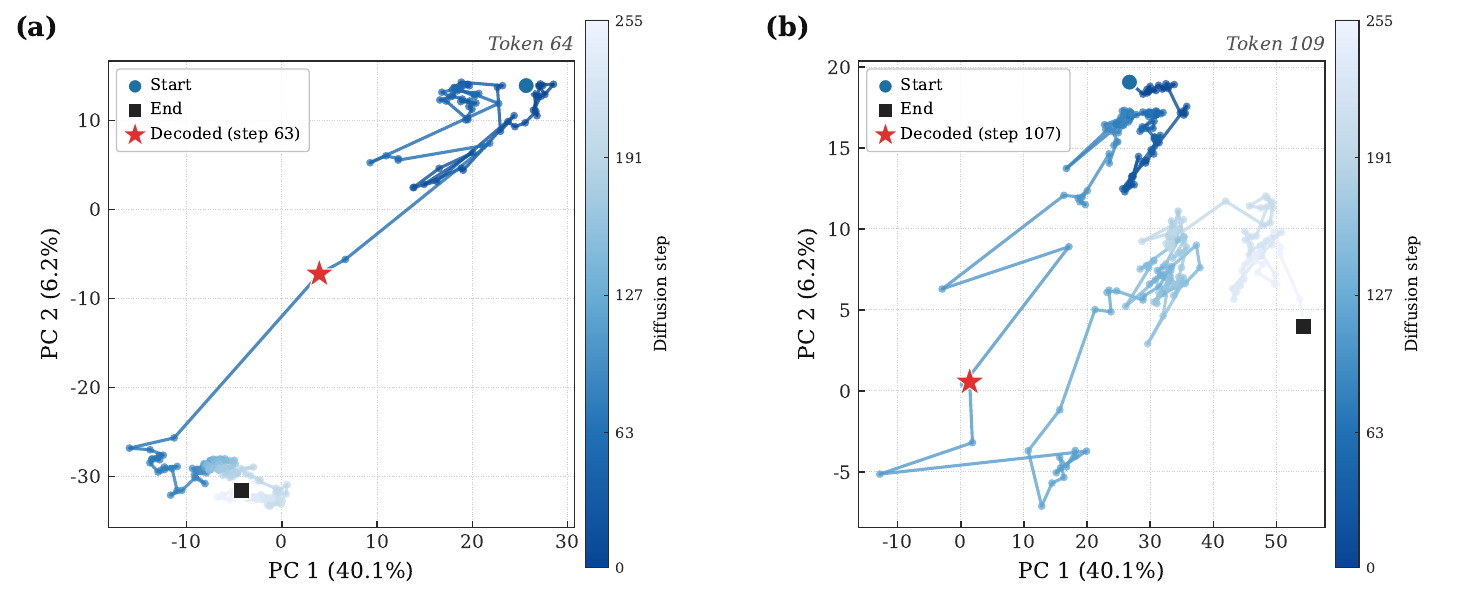}
    \caption{%
  (a)--(b)~PCA projections (PC\,1 vs.\ PC\,2) of the last-layer value vectors for two mask tokens over 256 denoising steps in LLaDA-8B-Instruct on a single GSM8K sample. Color encodes step progression (dark$\to$light); the red star marks the decoding step.
}
    \label{fig:figure2}
\end{figure}

The previous section established \emph{when} the KV cache drifts; we now examine \emph{which tokens} are most affected and for \emph{how long}. To this end, we track individual token trajectories in the value-vector space across all denoising steps.

\paragraph{PCA trajectories of individual tokens.}
Following $\text{d}^2\text{Cache}$~\citep{d2cache}, we apply PCA to the last-layer value vectors of all mask-token positions across 256 denoising steps and visualize two tokens in Figure~\ref{fig:figure2}(a)--(b). Token 64 experiences a sharp shift in trajectory directly after being decoded but converges rapidly in the following steps. By contrast, Token 109 continues to drift through the principal-component space for many steps after being decoded, exhibiting sustained volatility before eventual stabilization. These two cases illustrate a spectrum: while some tokens stabilize almost immediately upon unmasking, others remain volatile well beyond the single decode step.

\paragraph{Implication.}
Prior dynamic caching methods~\citep{d2cache,dkv-cache} acknowledge the importance of recomputing recently decoded tokens, yet restrict their recomputation only to the single following step. The above analysis suggests that this single-step scope may be insufficient, and that extending recomputation to a broader window of recently decoded tokens could better capture the feature dynamics.

\section{Method}
\label{sec:method}

\begin{figure}[ht!]
    \centering
    \includegraphics[width=\textwidth]{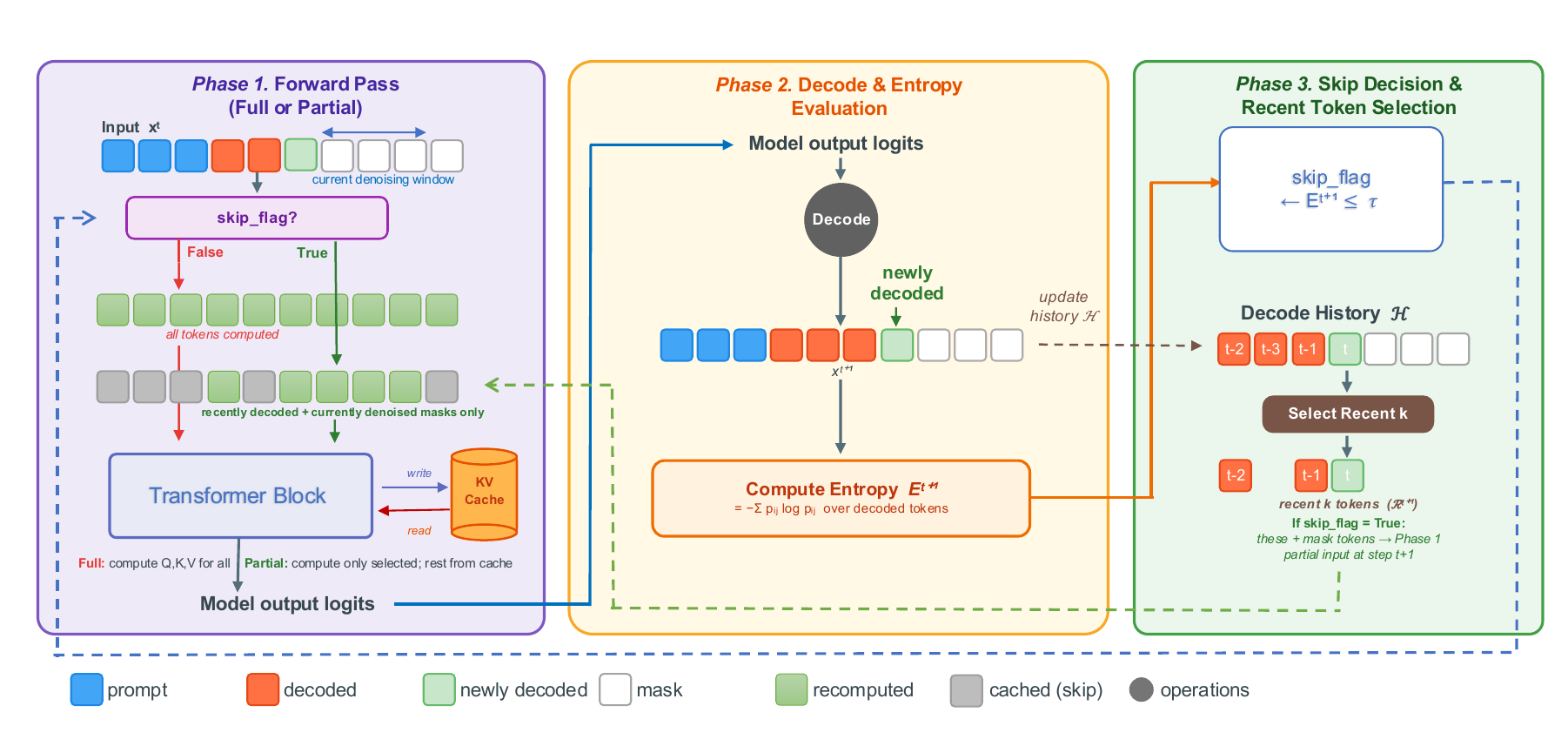}
    \caption{Overview of EntropyCache at a single denoising step $t$. \textbf{Phase~1}: a full or partial forward pass is executed depending on the recompute flag; in the partial case, only mask tokens and recently decoded tokens are recomputed while the rest are read from cache. \textbf{Phase~2}: new tokens are decoded from the model logits and the maximum entropy $E^{t+1}$ of the decoded distributions is computed. \textbf{Phase~3}: the entropy is compared against threshold $\tau$ to set the recompute flag for the next step, and the $k$ most recently decoded tokens $\mathcal{R}^{t+1}$ are selected for partial recomputation.}
    \label{fig:method_overview}
\end{figure}

Motivated by the observations in Section~\ref{sec:motivation}, we propose EntropyCache, a KV caching method for dLLMs that \textcircled{1} uses a single entropy scalar to decide \emph{when} to trigger full recomputation and \textcircled{2} recomputes the $k$ most recently decoded tokens to account for multi-step feature volatility. Figure~\ref{fig:method_overview} illustrates the three phases executed at each denoising step; we describe each component below.

\subsection{Entropy-based skipping}
\label{sec:entropy_skipping}

Section~\ref{sec:entropy_kv_drift} showed that the maximum entropy of newly decoded tokens is an effective predictor of KV cache drift. We exploit this by using it as a binary trigger for full recomputation.

At each step $t$, after decoding new tokens $\mathcal{D}^{t+1}$ from the model logits over all vocab indexes $j \in V$, we compute the maximum entropy over their predicted distributions:
\begin{equation}
    E^{t+1} = \max_{i \in \mathcal{D}^{t+1}} \Bigl(-\sum_j p_{ij} \log p_{ij}\Bigr).
\end{equation}
If $E^{t+1} > \tau$, the decoded tokens originate from uncertain distributions and a large cache drift is expected; therefore, we trigger a full forward pass at step $t{+}1$ to refresh the entire KV cache. If $E^{t+1} \leq \tau$, we reuse the existing KV cache, executing only a partial forward pass over the selected subset of tokens. Computing entropy for $N=|\mathcal{D}^{t+1}|$ newly decoded distributions costs $O(NV)$, where $V$ is the vocabulary size. This overhead is independent of context length, model depth, and hidden dimension, and $N$ is small under confidence-based parallel decoding.

\subsection{Recent token recomputation}
\label{sec:recent_tokens}

When the entropy-based decision skips full recomputation, we find that updating a small subset of recently decoded tokens alongside the current mask tokens is sufficient to retain model accuracy.
Section~\ref{sec:recent_token_volatility} demonstrates that feature volatility persists for multiple steps after a token is unmasked, not just the single step assumed by prior methods. Accordingly, we maintain a decode history $\mathcal{H} \in \mathbb{R}^L$ and at each step select a fixed budget of $k$ recently decoded tokens for recomputation.

Formally, each entry of the history vector records when a position was unmasked:
\begin{equation}
    \mathcal{H}_i = 
    \begin{cases}
        n & \text{if position } i \text{ was decoded at step } n, \\
        -\infty & \text{otherwise.}
    \end{cases}
\end{equation}
We then select the $k$ positions with the largest history values:
\begin{equation}
    \mathcal{K} = \underset{\mathcal{S} \subset \{1,\ldots,|\mathcal{H}|\},\; |\mathcal{S}|=k}{\arg\max} \sum_{j \in \mathcal{S}} \mathcal{H}_j,
\end{equation}
and define a recency threshold:
\begin{equation}
\tau_{\mathcal{R}} = \max\!\bigl(\min_{j \in \mathcal{K}} \mathcal{H}_j,\; t - \Delta t_{recompute}\bigr)
\end{equation}
where $\Delta t_{recompute}$ is the number of steps elapsed since the last full recomputation and $t$ is the current time step. Since a full forward pass resets all KV states to their exact values, subsequent drift is driven primarily by tokens decoded after that point; the threshold $\tau_{\mathcal{R}}$ therefore restricts the recomputation budget to these positions, where volatility is concentrated. The selected recent tokens mask is then $\mathcal{R}^{t+1} = \mathbb{I}(\mathcal{H}_i \geq \tau_{\mathcal{R}})$. In the partial forward pass, only the current mask tokens $\mathcal{M}^{t+1}$ and the selected recent tokens have their K, V recomputed; all other positions are reused from the cached KV states.

\paragraph{Why a fixed token budget rather than a fixed step window.}
An alternative design would recompute all tokens decoded within the last $n$ steps. However, when combined with confidence-based parallel decoding, the number of tokens unmasked per step varies substantially; high-confidence steps may decode dozens of tokens at once. A fixed step window therefore leads to unpredictable and often excessive recomputation in such steps, degrading throughput without proportional accuracy gains. By fixing the token budget $k$ instead, EntropyCache maintains a stable computational cost per partial step regardless of how many tokens are decoded at each iteration.

\subsection{Pseudocode}
\label{sec:pseudocode}

Algorithm~\ref{alg:pseudocode} summarizes the complete EntropyCache procedure. Starting from a full prefill of the prompt, each iteration decodes new tokens, evaluates entropy, and decides whether to perform a full or partial forward pass at the next step. The full algorithm with layer-level details is provided in Appendix~\ref{sec:full_algorithm}.

\SetKwComment{mycomment}{// }{}
\begin{algorithm}[ht!]
\small
\caption{EntropyCache}
\label{alg:pseudocode}
\SetAlgoLined
\DontPrintSemicolon
\SetAlgoNlRelativeSize{-1}
\setlength{\algomargin}{1.2em}
\KwIn{Prompt $\mathbf{x}_{\text{prompt}}$, generation length $N$, entropy threshold $\tau$, recent-token budget $k$}
\mycomment{Initialization}
$t \leftarrow 1$;\quad $\mathcal{H} \leftarrow \{-\infty\}^{N}$;\quad $\mathcal{A} \leftarrow \text{FullPrefill}(\mathbf{x}_{\text{prompt}})$;\quad $\mathcal{M}^1 \leftarrow \text{InitialWindow}$

\While{$\mathcal{M}^t \neq \emptyset$}{
    $\mathbf{x}^{t+1},\; \mathcal{D}^{t+1} \leftarrow \mathrm{decode}(\mathbf{x}^t,\, \mathcal{M}^t,\, \mathcal{A})$ \tcp*{Decode}
    $\mathcal{H}[\mathcal{D}^{t+1}] \leftarrow t$ \tcp*{Update history}
    $\mathcal{K} \leftarrow \underset{|S|=k}{\arg\max}\sum_{j\in S}\mathcal{H}_j$;\quad
    $\tau_{\mathcal{R}} \leftarrow \max\!\bigl(\min_{j\in\mathcal{K}}\mathcal{H}_j,\; t - \Delta t_{recompute}\bigr)$;\quad
    $\mathcal{R}^{t+1} \leftarrow \{i \mid \mathcal{H}_i \geq \tau_{\mathcal{R}}\}$\;
    $E^{t+1} \leftarrow \max_{i\in\mathcal{D}^{t+1}}\bigl(-\sum_j p_{ij}\log p_{ij}\bigr)$ \tcp*{Entropy}
    \eIf{$E^{t+1} \leq \tau$}{
        $\mathcal{A} \leftarrow \mathcal{A}$ \tcp*{reuse KV-cache with $\mathcal{R}^{t+1}$}
    }{
        $\mathcal{A} \leftarrow \mathrm{FullRecompute}(\mathbf{x}^{t+1})$\;
    }
    $\mathcal{M}^{t+1} \leftarrow \mathcal{M}^t \setminus \mathcal{D}^{t+1}$;\quad $t \leftarrow t+1$\;
}
\Return{$\mathbf{x}^{t}$}
\end{algorithm}

\section{Experiments}

\subsection{Experimental setup}

Unless otherwise specified, all experiments are conducted on NVIDIA RTX 3090 GPUs. To ensure a fair comparison of
KV cache mechanisms, we re-implement all methods within a unified codebase to share the
same decoding logic. We evaluate LLaDA-8B-Instruct and Dream-7B-Instruct using sliding
window decoding (window size = 32), except for Fast-dLLM, which strictly requires blockwise
decoding (block size = 32). All KV cache methods use confidence-based parallel
decoding with a threshold of 0.9. For dynamic KV cache methods, we focus on d$^2$Cache and Elastic-Cache as baselines, as prior work~\cite{d2cache, elasticcache} has shown them to be more effective than other previous approaches~\cite{dkv-cache, dllm-cache} under comparable settings. We use the optimal
parameters reported in their respective papers: $\text{rollout}_p = 0.2$ for $\text{d}^2\text{Cache}$ and a
cosine similarity threshold of 0.9 for Elastic-Cache. For EntropyCache, we set $\tau = 1.5$
and $k = 64$.

We evaluate on two benchmark suites: (1)~standard benchmarks including
GSM8K~\citep{gsm8k} 4-shot, MATH500~\citep{math-500} 4-shot,
MBPP~\citep{mbpp} 3-shot, and HumanEval~\citep{humaneval} zero-shot; and
(2)~chain-of-thought benchmarks including GSM8K 8-shot, MMLU-Pro~\citep{mmlu-pro},
BBH 3-shot~\citep{bbh}, and GPQA zero-shot~\citep{gpqa}.
\subsection{Main results}
\begin{table*}[!ht]
\centering
\caption{Comprehensive dLLM KV cache method comparison. Speedup ($\times$) relative to Baseline. \underline{Underlined} scores meet or exceed Baseline accuracy. Acc.: Accuracy (\%). T-put: Throughput (tok/s). Superscripts indicate max generation length.}
\label{tab:main_result_basic}
\resizebox{\textwidth}{!}{%
\begin{tabular}{@{}l*{10}{c}@{}}
\toprule
& \multicolumn{2}{c}{\textbf{GSM8K}$^{256}$} & \multicolumn{2}{c}{\textbf{MATH500}$^{512}$} & \multicolumn{2}{c}{\textbf{MBPP}$^{256}$} & \multicolumn{2}{c}{\textbf{HumanEval}$^{512}$} & \multicolumn{2}{c}{\textbf{Average}} \\
\cmidrule(lr){2-3} \cmidrule(lr){4-5} \cmidrule(lr){6-7} \cmidrule(lr){8-9} \cmidrule(lr){10-11}
\textbf{Method} & Acc. & T-put & Acc. & T-put & Acc. & T-put & Acc. & T-put & Acc. & T-put \\
\midrule
\multicolumn{11}{c}{\textit{LLADA-8B-Instruct}} \\
\midrule
Baseline               & 77.86 & 2.28  & 42.40 & 2.47  & 42.20 & 2.89  & 46.34 & 4.32  & 52.20 & 2.99  \\
Baseline+Parallel      & \underline{78.47} & 8.56\,(3.8$\times$)   & \underline{42.40} & 9.50\,(3.9$\times$)   & \underline{42.40} & 16.37\,(5.7$\times$)  & \underline{48.17} & 15.40\,(3.6$\times$)  & \underline{52.86} & 12.46\,(4.2$\times$)  \\
\midrule[0.01pt]
Fast-DLLM (Dual)       & \underline{78.01} & 32.54\,(14.3$\times$) & 40.00 & 32.81\,(13.3$\times$) & 39.20 & 45.58\,(15.8$\times$) & 44.51 & 37.71\,(8.7$\times$)  & 50.43 & 37.16\,(12.4$\times$) \\
Elastic-Cache          & \underline{79.15} & 23.49\,(10.3$\times$) & \underline{43.80} & 24.43\,(9.9$\times$)  & 40.60 & 32.24\,(11.2$\times$) & 44.51 & 27.73\,(6.4$\times$)  & 52.02 & 26.97\,(9.0$\times$)  \\
$\text{d}^2\text{Cache}$ & \underline{78.01} & 29.96\,(13.2$\times$) & \underline{43.60} & 32.18\,(13.1$\times$) & 40.80 & 52.77\,(18.3$\times$) & 41.46 & 41.36\,(9.6$\times$)  & 50.97 & 39.06\,(13.1$\times$) \\
\rowcolor{gray!10}
EntropyCache (Ours)    & \underline{78.77} & 38.49\,(16.9$\times$) & \underline{42.80} & 39.64\,(16.1$\times$) & 38.80 & 64.91\,(22.5$\times$) & \underline{48.78} & 38.41\,(8.9$\times$)  & \underline{\textbf{52.28}} & \textbf{45.36}\,$\bm{(15.2\times)}$ \\
\midrule
\multicolumn{11}{c}{\textit{Dream-v0-Instruct-7B}} \\
\midrule
Baseline               & 74.30 & 2.60  & 44.60 & 2.60  & 50.60 & 3.31  & 53.05 & 3.87  & 55.64 & 3.10  \\
Baseline+Parallel      & \underline{74.91} & 12.14\,(4.7$\times$)  & 44.00 & 15.86\,(6.1$\times$)  & 48.60 & 29.66\,(9.0$\times$)  & \underline{53.05} & 38.74\,(10.0$\times$) & 55.14 & 24.10\,(7.8$\times$)  \\
\midrule[0.01pt]
Fast-DLLM (Dual)       & 72.25 & 42.17\,(16.2$\times$) & 40.20 & 43.52\,(16.7$\times$) & \underline{51.40} & 69.36\,(21.0$\times$) & \underline{55.49} & 71.71\,(18.5$\times$) & 54.83 & 56.69\,(18.3$\times$) \\
Elastic-Cache          & 69.60 & 32.15\,(12.4$\times$) & 38.80 & 41.90\,(16.1$\times$) & 45.80 & 54.84\,(16.6$\times$) & 50.61 & 72.30\,(18.7$\times$) & 51.20 & 50.30\,(16.2$\times$) \\
$\text{d}^2\text{Cache}$ & \underline{75.44} & 41.68\,(16.0$\times$) & 40.60 & 55.80\,(21.4$\times$) & 48.40 & 87.63\,(26.5$\times$) & \underline{53.66} & 110.83\,(28.6$\times$) & 54.52 & 73.98\,(23.9$\times$) \\
\rowcolor{gray!10}
EntropyCache (Ours)    & \underline{74.30} & 48.32\,(18.6$\times$) & 43.20 & 62.48\,(24.0$\times$) & 48.80 & 97.79\,(29.6$\times$) & \underline{58.54} & 119.02\,(30.7$\times$) & \underline{\textbf{56.21}} & \textbf{81.90}\,$\bm{(26.4\times)}$ \\
\bottomrule
\end{tabular}%
}
\end{table*}
\begin{table*}[!ht]
\centering
\caption{Chain-of-Thought (CoT) benchmark comparison. Speedup ($\times$) relative to Baseline. Acc.: Accuracy (\%). T-put: Throughput (tok/s).}
\label{tab:cot_benchmarks}
\resizebox{\textwidth}{!}{%
\begin{tabular}{@{}l*{10}{c}@{}}
\toprule
& \multicolumn{2}{c}{\textbf{GSM8K 8-shot}} & \multicolumn{2}{c}{\textbf{MMLU-Pro}} & \multicolumn{2}{c}{\textbf{BBH 3-shot}} & \multicolumn{2}{c}{\textbf{GPQA Zero-shot}} & \multicolumn{2}{c}{\textbf{Average}} \\
\cmidrule(lr){2-3} \cmidrule(lr){4-5} \cmidrule(lr){6-7} \cmidrule(lr){8-9} \cmidrule(lr){10-11}
\textbf{Method} & Acc. & T-put & Acc. & T-put & Acc. & T-put & Acc. & T-put & Acc. & T-put \\
\midrule
\multicolumn{11}{c}{\textit{LLADA-8B-Instruct}} \\
\midrule
Baseline               & 81.12 & 2.07  & 37.60 & 1.68  & 59.58 & 1.35   & 27.68 & 5.76  & 51.49 & 2.71  \\
Baseline+Parallel      & 81.43 & 12.44\,(6.0$\times$)   & 37.57 & 4.77\,(2.8$\times$)   & 57.90 & 21.45\,(15.9$\times$)  & 27.68 & 11.22\,(1.9$\times$)  & 51.15 & 12.47\,(4.6$\times$)  \\
\midrule[0.01pt]
Fast-DLLM (Dual)       & 80.44 & 41.68\,(20.1$\times$)  & 37.15 & 21.56\,(12.8$\times$)  & 52.48 & 36.89\,(27.3$\times$)  & 24.78 & 25.34\,(4.4$\times$)  & 48.71 & 31.37\,(11.6$\times$)  \\
Elastic-Cache          & 81.20 & 32.89\,(15.9$\times$)  & 36.69 & 14.78\,(8.8$\times$)   & 52.57 & 42.39\,(31.4$\times$)  & 24.78 & 18.08\,(3.1$\times$)  & 48.81 & 27.04\,(10.0$\times$)  \\
$\text{d}^2\text{Cache}$ & 81.35 & 47.03\,(22.7$\times$)  & 36.95 & 16.80\,(10.0$\times$)  & 55.68 & 83.94\,(62.2$\times$)  & 22.10 & 23.22\,(4.0$\times$)  & \textbf{49.02} & 42.75\,(15.8$\times$)  \\
\rowcolor{gray!10}
EntropyCache (Ours)    & 81.20 & 62.31\,(30.1$\times$) & 36.06 & 15.21\,(9.1$\times$)   & 52.48 & 144.90\,(107.3$\times$) & 24.55 & 20.98\,(3.6$\times$)  & 48.57 & \textbf{60.85}\,(22.4$\times$)  \\
\midrule
\multicolumn{11}{c}{\textit{Dream-v0-Instruct-7B}} \\
\midrule
Baseline               & 78.01 & 2.51  & 47.12 & 2.01  & 65.69 & 1.61   & 27.01 & 6.14  & 54.46 & 3.07  \\
Baseline+Parallel      & 77.41 & 13.95\,(5.6$\times$)   & 46.64 & 10.93\,(5.4$\times$)   & 60.76 & 25.42\,(15.8$\times$)  & 24.78 & 19.12\,(3.1$\times$)  & 52.40 & 17.36\,(5.7$\times$)  \\
\midrule[0.01pt]
Fast-DLLM (Dual)       & 74.22 & 55.50\,(22.1$\times$)  & 41.86 & 35.01\,(17.4$\times$)  & 58.09 & 40.96\,(25.4$\times$)  & 25.22 & 34.69\,(5.6$\times$)  & 49.85 & 41.54\,(13.5$\times$)  \\
Elastic-Cache          & 69.37 & 36.95\,(14.7$\times$)  & 42.50 & 26.50\,(13.2$\times$)  & 31.52 & 52.61\,(32.7$\times$)  & 27.68 & 26.90\,(4.4$\times$)  & 42.77 & 35.74\,(11.6$\times$)  \\
$\text{d}^2\text{Cache}$ & 73.69 & 49.59\,(19.8$\times$)  & 43.33 & 40.95\,(20.4$\times$)  & 59.51 & 99.60\,(61.8$\times$)  & 27.01 & 35.23\,(5.7$\times$)  & 50.89 & 56.34\,(18.4$\times$)  \\
\rowcolor{gray!10}
EntropyCache (Ours)    & 75.74 & 60.01\,(23.9$\times$)  & 45.49 & 42.14\,(21.0$\times$)  & 55.86 & 155.30\,(96.4$\times$) & 26.79 & 38.73\,(6.3$\times$)  & \textbf{50.97} & \textbf{74.05}\,$\bm{(24.1\times)}$ \\
\bottomrule
\end{tabular}%
}
\end{table*}

\paragraph{Standard benchmarks.}
Table~\ref{tab:main_result_basic} compares all methods on four standard benchmarks.
EntropyCache achieves the best accuracy and throughput among all caching methods:
on LLaDA-Instruct it attains 52.28\% average accuracy
at $15.2\times$ speedup, and on Dream-Instruct it reaches 56.21\% at $26.4\times$ speedup.
No other caching method preserves baseline-level average accuracy while achieving comparable
throughput. The advantage is clearest on HumanEval, where all other methods degrade accuracy on at least one
model.
EntropyCache is the only method that on average achieves lossless performance compared to baseline on both models, while still delivering $8.9\times$--$30.7\times$ speedup.

\paragraph{Chain-of-thought benchmarks.}
Table~\ref{tab:cot_benchmarks} evaluates on longer-context CoT tasks.
EntropyCache achieves $22.4\times$ and $24.1\times$ average speedup on LLaDA-Instruct and Dream-Instruct, roughly $1.4\times$ the next-best method ($\text{d}^2\text{Cache}$).
On BBH 3-shot the gap widens to $1.7\times$, with EntropyCache reaching $107.3\times$ speedup on LLaDA and $96.4\times$ on Dream, as long generations with many low-entropy steps maximize cache reuse.
Accuracy is highest among caching methods on Dream-Instruct (50.97\%) and within 0.5pp of $\text{d}^2\text{Cache}$ on LLaDA-Instruct (48.57\% vs.\ 49.02\%).

EntropyCache's performance on BBH and MBPP is further analyzed in Section~\ref{sec:ablation} and Appendix~\ref{app:extended_ablation}, respectively: on BBH, EntropyCache Pareto-dominates $\text{d}^2\text{Cache}$ once both methods are swept across configurations, and on MBPP the accuracy is largely preserved across a range of $\tau$ values, indicating that the default-configuration drop reflects a local minimum rather than a fundamental limitation.

\subsection{Ablation study}
\label{sec:ablation}

\begin{figure}[ht!]  
    \centering
    \includegraphics[width=\linewidth]{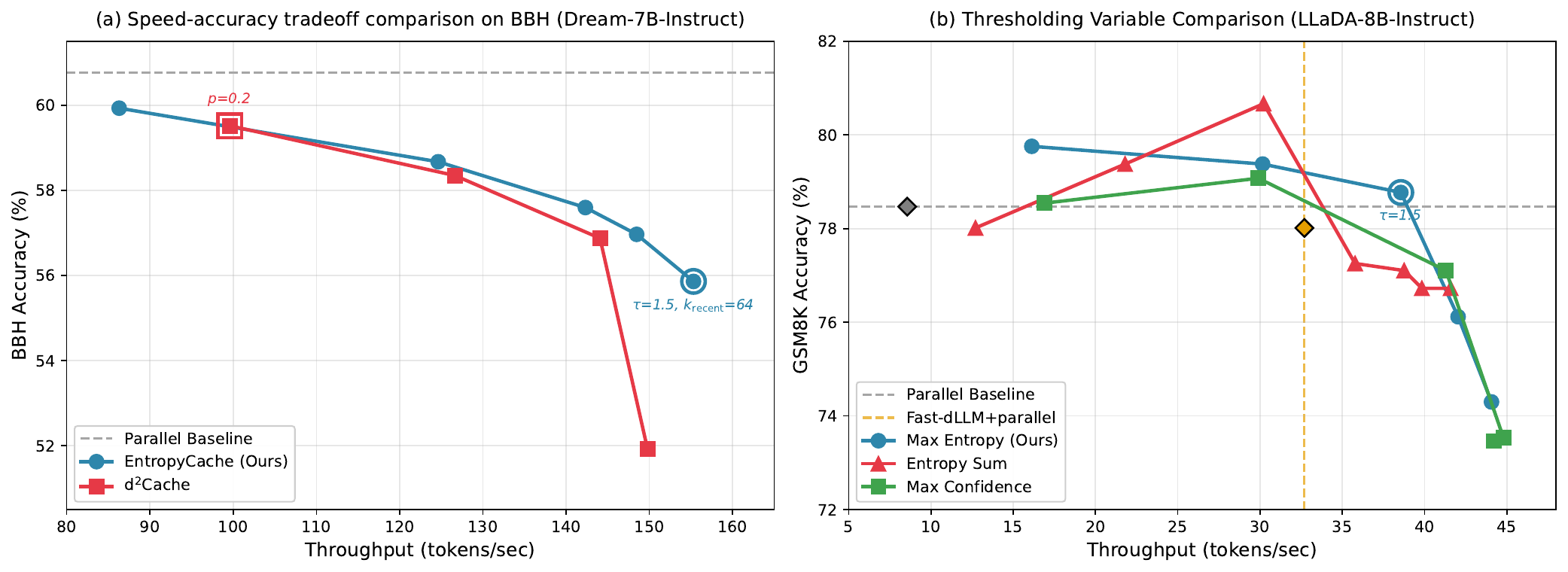}
    \caption{
        \textbf{(a)} Speed-accuracy tradeoff comparison on BBH between EntropyCache and $\text{d}^2\text{Cache}$ (Dream-7B-Instruct). Hollow markers denote default configurations. 
        \textbf{(b)} Accuracy--throughput tradeoff on GSM8K (LLaDA-8B-Instruct) for three candidate skipping metrics across varying thresholds $\tau$.
    }
    \label{fig:ablation_combined}
\end{figure}

\paragraph{Speed-accuracy tradeoff comparison on BBH.}
We further analyze EntropyCache's behavior on long-generation tasks by tracing out the speed-accuracy tradeoff on BBH with Dream-7B-Instruct, going beyond the single-configuration view of Table~\ref{tab:cot_benchmarks}.
We sweep $\tau$ for EntropyCache and $\mathrm{rollout}_p$ for d$^2$Cache (Figure~\ref{fig:ablation_combined}(a)); the default configurations of the two methods land at different points on the tradeoff curve, which makes a single-configuration comparison misleading.
Across the full sweep, EntropyCache consistently outperforms $\text{d}^2\text{Cache}$: at matched throughput, EntropyCache achieves higher accuracy.
EntropyCache also maintains stable accuracy at high throughput, whereas d$^2$Cache degrades sharply as $\mathrm{rollout}_p$ decreases.

\paragraph{Choice of thresholding variable}
We compare three candidate metrics for adaptive skipping:
(1)~max entropy, which takes the maximum entropy of decoded tokens;
(2)~entropy sum, which uses the summation of entropy over all decoded tokens;
and (3)~max confidence, which uses the maximum token-level probability across decoded tokens.
For each metric, we sweep the threshold~$\tau$ and plot the resulting accuracy--throughput tradeoff on GSM8K (Figure~\ref{fig:ablation_combined}(b)).
Max entropy produces the most favorable tradeoff among the three: at $\tau=1.5$ it achieves approximately 78.8\% accuracy at ${\sim}$38.6~tok/s, surpassing both the parallel baseline's accuracy and Fast-dLLM DualCache's throughput.
Entropy sum peaks sharply at $\tau=1.5$ but degrades beyond it, making threshold selection sensitive to miscalibration.
Max confidence is dominated by max entropy across most of the throughput range.
Based on these results, we adopt max entropy as the default thresholding variable for all experiments.

\begin{figure}[ht!]
    \centering
    \includegraphics[width=\linewidth]{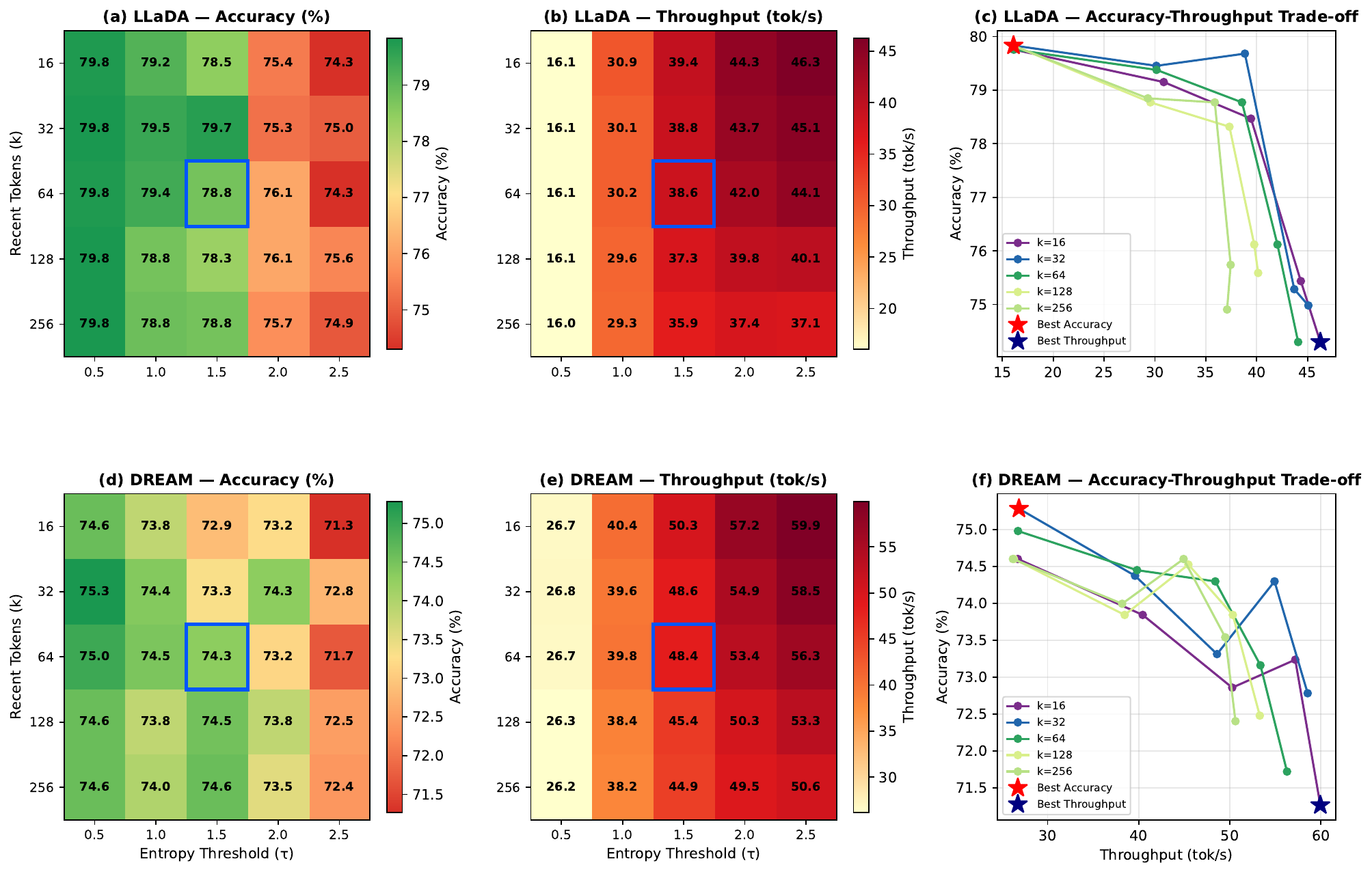}
    \caption{Joint $\tau$--$k_\mathrm{recent}$ sweep on GSM8K. Top: LLaDA; bottom: Dream. Left to right: accuracy, throughput, and their trade-off.}
    \label{fig:grid_search}
\end{figure}

\paragraph{Joint parameter sensitivity.}
Figure~\ref{fig:grid_search} jointly sweeps $\tau$ and $k_\mathrm{recent}$ on GSM8K for both models. The entropy threshold controls the primary accuracy--throughput trade-off, while the recent-token budget provides complementary accuracy recovery at modest cost. The shared setting $\tau=1.5$, $k_\mathrm{recent}=64$ lies near the knee for both models; on Dream, increasing $k_\mathrm{recent}$ from 16 to 64 improves accuracy from 72.9\% to 74.3\% for less than 2 tok/s.

\paragraph{Robustness across hyperparameters and inference settings.}
Appendix~\ref{app:extended_ablation} evaluates robustness to the sliding-window size and reports per-benchmark $\tau$--$k_\mathrm{recent}$ sweeps across all four standard benchmarks. Appendix~\ref{app:additional_validation} provides matched operating-point comparisons and deployment-oriented analyses.

\subsection{Overhead analysis}

\begin{figure}[ht!]
    \centering
    \includegraphics[width=\textwidth]{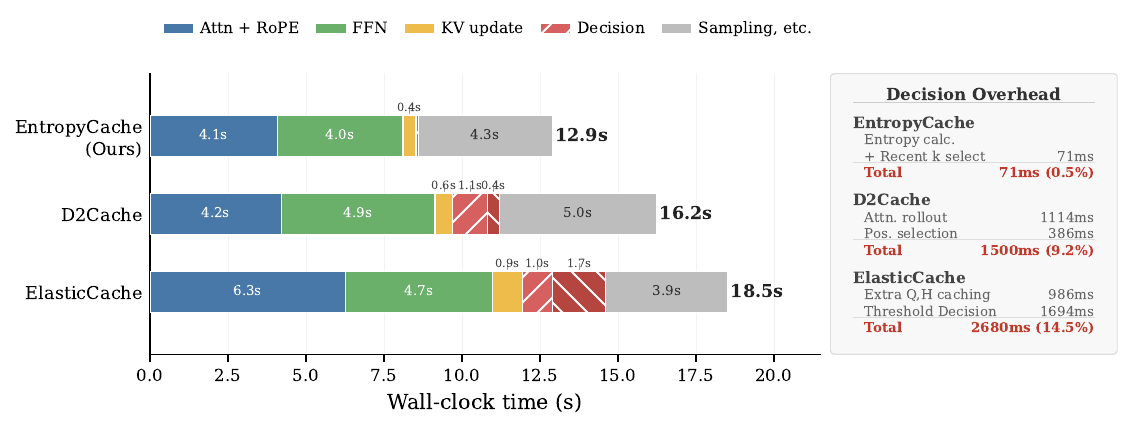}
    \caption{Wall-clock time breakdown on a single GSM8K sample (LLaDA-Inst, $w = 32$, generation length $= 256$). Left: stacked bar chart of inference components. Right: decision overhead decomposition per method. EntropyCache's decision cost is 71\,ms (0.5\%), roughly $20\times$ and $38\times$ cheaper than $\text{d}^2\text{Cache}$ and Elastic-Cache respectively.}
    \label{fig:overhead_breakdown}
\end{figure}

Dynamic KV caching methods introduce extra computation at each step to decide where or when to refresh the cache.
A key practical advantage of EntropyCache is that its decision computation is decoupled from both context length and model scale.
Table~\ref{tab:overhead} compares the per-step complexity of the three dynamic caching methods. 
$\text{d}^2\text{Cache}$ computes attention rollout across all layers and heads, incurring $O(\ell \cdot L \cdot H)$ computation and $O(H \cdot L^2)$ memory, quadratic in context length. 
Elastic-Cache performs layer-wise partial attention score calculation at cost $O(\ell \cdot L \cdot d)$, scaling with model depth, context length, and hidden dimension simultaneously. 
In contrast, EntropyCache's skip decision only requires entropy evaluation over the $N$ newly decoded vocabulary distributions, costing $O(NV)$ computation, plus $O(T)$ memory for the decode history over $T$ generated positions.
Although this cost is asymptotically linear in $V$, it is entirely independent of context length $L$, model depth $\ell$, and hidden dimension $d$; moreover, confidence-based decoding typically yields only a few newly decoded tokens per step. Appendix~\ref{app:vocab_scaling} quantifies this cost across vocabulary sizes from 32K to 320K.
Wall-clock profiling confirms this gap: EntropyCache's decision logic accounts for only 0.5\% of inference time, versus 9.2\% for $\text{d}^2\text{Cache}$ and 14.5\% for Elastic-Cache (Figure~\ref{fig:overhead_breakdown}).


\begin{table}[!ht]
  \centering
  \caption{%
    Complexity comparison of dynamic KV cache methods.
    EntropyCache decouples decision computation
    from both context length and model scale.
    \newline
    \textit{Notation:}
    $\ell$ = number of layers;
    $L$ = context length;
    $H$ = number of attention heads;
    $d$ = hidden dimension;
    $V$ = vocabulary size;
    $N$ = number of newly decoded tokens;
    $T$ = number of generated positions.
    $\text{d}^2\text{Cache}$ and Elastic-Cache overhead grows with $L$ and model scale
    ($\ell$, $d$), respectively, whereas EntropyCache's decision computation is independent
    of both context length and model scale.
  }
  \label{tab:overhead}
  \setlength{\tabcolsep}{7pt}       
  \renewcommand{\arraystretch}{1} 
  %
  \begin{tabular}{@{} l c c l @{}}
    \toprule
    \textbf{Method}
      & \textbf{Computation}
      & \textbf{Memory}
      & \textbf{Bottleneck} \\
    \midrule
    $\text{d}^2\text{Cache}$~\cite{d2cache}
      & $\mathcal{O}(\ell \cdot L \cdot H)$
      & $\mathcal{O}(H \cdot L^{2})$
      & Quadratic in $L$ \\
    Elastic-Cache~\cite{elasticcache}
      & $\mathcal{O}(\ell \cdot L \cdot d)$
      & $\mathcal{O}(\ell \cdot L \cdot d)$
      & Scales with $\ell$, $L$, $d$ \\
    \textbf{EntropyCache (Ours)}
      & $\mathcal{O}(N \cdot V)$
      & $\mathcal{O}(T)$
      & \textbf{Compute independent of $L,\ell,d$} \\
    \bottomrule
  \end{tabular}
  %
  \smallskip
  \begin{minipage}{0.92\linewidth}
    \footnotesize
    
  \end{minipage}
\end{table}

\section{Conclusion}

We presented EntropyCache, a KV caching method for diffusion-based large language models that provides a cheap proxy for deciding \emph{when} recomputation is necessary. By exploiting the empirical correlation between decoded token entropy and KV cache drift, EntropyCache uses a lightweight $O(NV)$ entropy check over newly decoded token distributions to decide whether to reuse cached states or trigger a full forward pass, while recomputing the $k$ most recently decoded tokens to account for multi-step feature volatility. On standard benchmarks, EntropyCache achieves 15.2$\times$ to 26.4$\times$ average speedup over vanilla dLLM inference with competitive accuracy across all baselines, and maintains consistent gains on chain-of-thought benchmarks that stress longer contexts. Additional ablation studies, overhead profiling, and extended empirical analyses are provided in the appendix.

\clearpage

\begin{ack}
This work was supported by Samsung Electronics (MX Business), IITP/NRF grants funded by the Korean government (MSIT, 2021-0-00105 Development of Model Compression Framework for Scalable On-Device AI Computing on Edge Applications, 2021-0-01343 Artificial Intelligence Graduate School Program, Seoul National University), and Inter-university Semiconductor Research Center (ISRC), SNU.
The authors declare no competing interests.
\end{ack}

\bibliography{ref}
\bibliographystyle{plain}

\newpage
\appendix

\section{Full EntropyCache algorithm}
\label{sec:full_algorithm}

\begin{algorithm}[H]
\caption{EntropyCache algorithm}
\SetAlgoLined
\DontPrintSemicolon

\KwIn{Prompt $\mathbf{x}_{\text{prompt}}$, Generation length $N$, Window size $w$, Entropy threshold $\tau$, Recent token window $k$}
\KwData{$\mathbf{x}^0 \leftarrow \{\mathbf{x}_{\text{prompt}}; \text{[MASK]}, \ldots, \text{[MASK]}\}$; \quad $P \leftarrow \text{length}(\mathbf{x}_{\text{prompt}})$}

$t \leftarrow 1$; \quad $\mathcal{D}^1 \leftarrow \{1, \ldots, P\}$; \quad $\mathcal{M}^1 \leftarrow \{P+1, \ldots, P+N\}$; \quad $\mathcal{R}^1 \leftarrow \emptyset$;

$\mathcal{A}^{0,l} \leftarrow \emptyset$ for all layers $l$; \quad $\mathcal{D}_{\text{prev}} \leftarrow \emptyset$; \quad $\text{skip\_flag} \leftarrow \text{False}$\;
$\mathcal{H} \leftarrow \{-\infty\}^N$\;

\While{$\mathcal{M}^t \neq \emptyset$}{
    \uIf{$t == 1 \text{ or } \text{skip\_flag} = \text{False}$}{
        \tcp{Prefill}
        \For{$l = 1, \ldots, L$}{
            $\mathbf{Q}_{[\mathcal{I}]}^{t,l}, \mathbf{K}_{[\mathcal{I}]}^{t,l}, \mathbf{V}_{[\mathcal{I}]}^{t,l}
            \leftarrow \text{W}(\mathbf{H}_{[\mathcal{I}]}^{t,l-1})$\;
            $\mathbf{H}_{[\mathcal{I}]}^{t,l}, \mathbf{S}_{[\mathcal{I}]} \leftarrow \text{TransformerBlock}
            (\mathbf{Q}_{[\mathcal{I}]}^{t,l} , \mathbf{K}_{[\mathcal{I}]}^{t,l}, \mathbf{V}_{[\mathcal{I}]}^{t,l})$\;
        }
    }\Else{
        \For{$l = 1, \ldots, L$}{
            
            $\mathbf{H}_{[\mathcal{M}^t]}^{t,0} \leftarrow \mathbf{x}_{[\mathcal{M}^t]}^t$\;
            
            $\mathbf{Q}_{[\mathcal{M}^t \cup \mathcal{R}^t]}^{t,l}, \mathbf{K}_{[\mathcal{M}^t \cup \mathcal{R}^t]}^{t,l}, \mathbf{V}_{[\mathcal{M}^t \cup \mathcal{R}^t]}^{t,l} \leftarrow \text{W}(\mathbf{H}_{[\mathcal{M}^t \cup \mathcal{R}^t]}^{t,l-1})$\;
            
            $\mathbf{H}_{[\mathcal{M}^t \cup \mathcal{R}^t]}^{t,l}, \mathbf{S}_{[\mathcal{M}^t \cup \mathcal{R}^t]}^{t,l} \leftarrow \text{TransformerBlock}(\mathbf{Q}_{[\mathcal{M}^t \cup \mathcal{R}^t]}^{t,l} , \mathbf{K}_{[\mathcal{I}]}^{t,l}, \mathbf{V}_{[\mathcal{I}]}^{t,l})$\;
            
        }
    }
    
    Decode new tokens: $\mathbf{x}^{t+1}, \mathcal{D}^{t+1} \leftarrow \text{decode}(\mathbf{x}^t, \mathcal{M}^t)$\;

    Record Decode history: $\mathcal{H}[\mathcal{D}^{t+1}] \leftarrow t$\;

    \tcp{Select recent tokens}
    $\mathcal{K} \leftarrow \underset{S \subset \{1,\ldots,|\mathcal{H}|\},\, |S|=k}{\arg\max} \displaystyle\sum_{j \in S} \mathcal{H}_j$\;
    $\tau_{\mathcal{R}} \leftarrow \max\!\left(\min_{j \in \mathcal{K}} \mathcal{H}_j,\; t - \Delta t_{recompute}\right)$\;
    $\mathcal{R}^{t+1} \leftarrow \{i \mid \mathcal{H}_i \geq \tau_{\mathcal{R}}\}$\;

    $E^{t+1} \leftarrow \max_{i\in\mathcal{D}^{t+1}}\bigl(e(p_i))$ where $e(p_i) = -\sum_j p_{ij} \log p_{ij}$ \tcp{Compute entropy for next step}
    
    \eIf{$E^{t+1} \leq \tau$}{
        $\text{skip\_flag} \leftarrow \text{True}$ \tcp{Set flag for next iteration}
        $\Delta t_{recompute} \leftarrow \Delta t_{recompute} + 1$\;
    }{
        $\text{skip\_flag} \leftarrow \text{False}$\;
        $\Delta t_{recompute} \leftarrow 0$\;
    }
    
    Update mask positions: $\mathcal{M}^{t+1} \leftarrow \mathcal{M}^t \setminus \mathcal{D}^{t+1}$\;
    
    $t \leftarrow t+1$\;
}

\Return{$\mathbf{x}^{t-1}$}
\end{algorithm}

\section{Extended empirical analysis on entropy and KV cache drift}
\label{app:extended_entropy_analysis}

\begin{figure}[!ht]
    \centering
    \includegraphics[width=\textwidth]{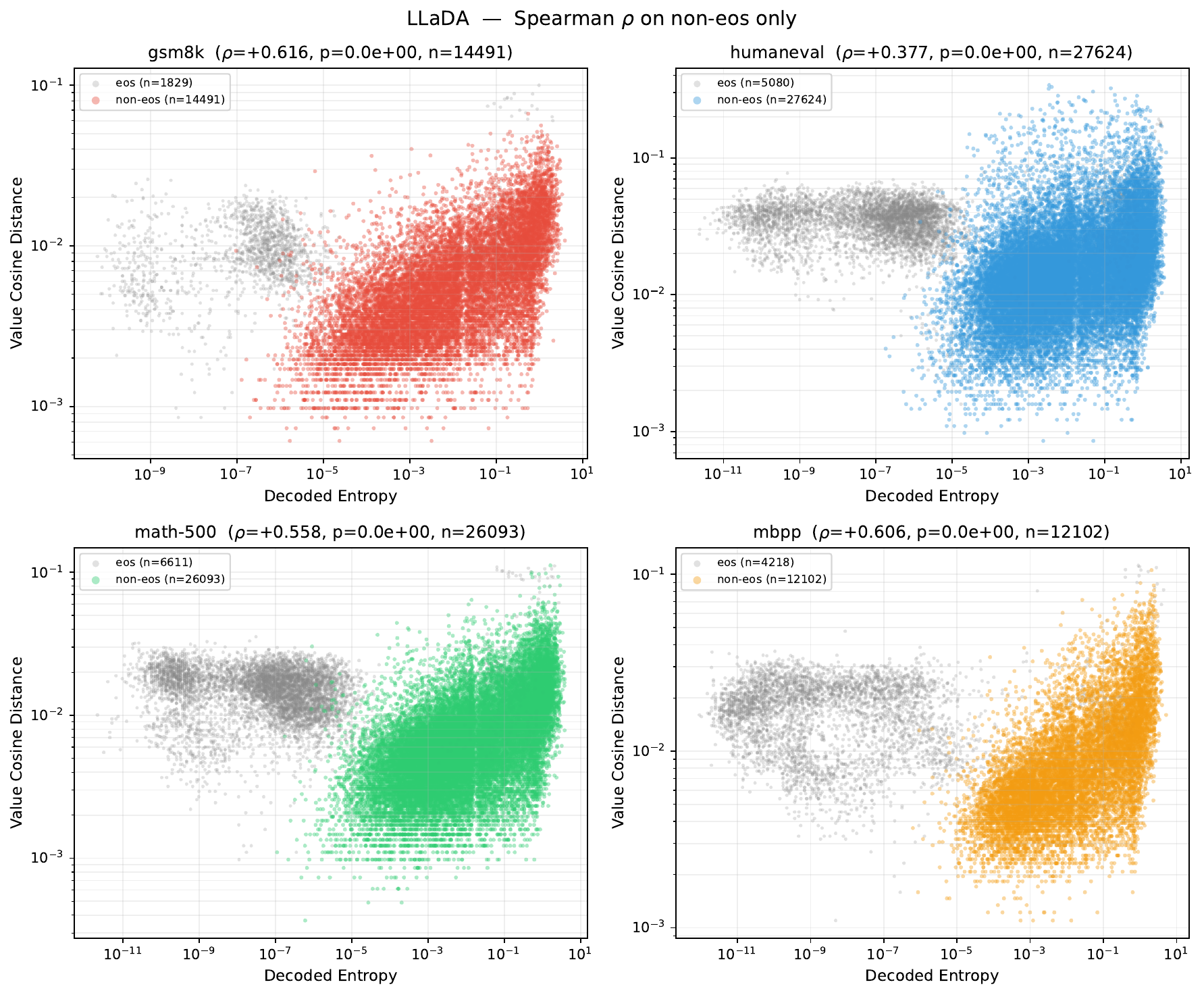}
    \caption{Per-step decoded token entropy vs.\ average value-vector cosine distance for each benchmark (LLaDA-8B-Instruct, 64 samples per dataset). Gray points denote EOS-filling steps; colored points denote non-EOS content steps. Spearman $\rho$ is computed on non-EOS steps only.}
    \label{fig:scatter_per_dataset}
\end{figure}

\begin{figure}[!ht]
    \centering
    \includegraphics[width=0.8\textwidth]{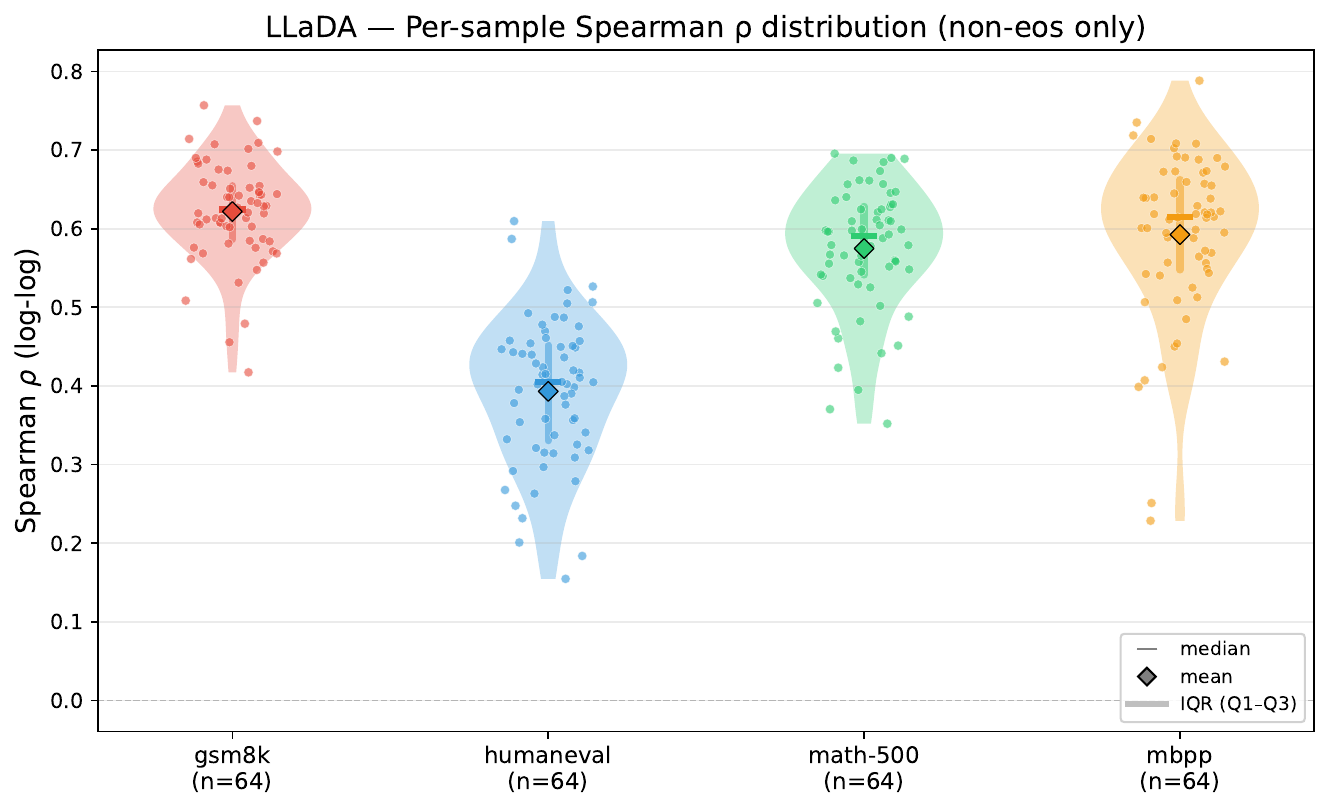}
    \caption{Distribution of per-sample Spearman $\rho$ (log--log, non-EOS only) across 64 samples for each benchmark (LLaDA-8B-Instruct). Diamonds indicate means; horizontal bars indicate medians; shaded regions show kernel density estimates.}
    \label{fig:spearman_distribution}
\end{figure}

The correlation between decoded token entropy and KV cache drift presented in Section~\ref{sec:motivation} was derived from a single generation trajectory. To validate that this relationship holds broadly, we extend the analysis across all four standard benchmarks (GSM8K, MATH500, MBPP, HumanEval), generating 64 samples per dataset using LLaDA-8B-Instruct under the same sliding-window decoding configuration used in our main experiments. For each denoising step of every sample, we record the maximum decoded token entropy and the average value-vector cosine distance across all token positions, yielding between 12K and 28K step-level observations per dataset.

\paragraph{Per-dataset scatter analysis.}
Figure~\ref{fig:scatter_per_dataset} plots the per-step entropy--drift pairs on log--log axes for each benchmark. Across all four datasets, the positive monotonic trend observed in Section~\ref{sec:entropy_kv_drift} is clearly reproduced: higher decoded entropy at step $t$ is consistently associated with larger KV cache drift at step $t{+}1$. The aggregate Spearman correlations on non-EOS steps are $\rho = 0.616$ (GSM8K), $\rho = 0.558$ (MATH500), $\rho = 0.606$ (MBPP), and $\rho = 0.377$ (HumanEval), all with $p < 0.001$.

\paragraph{EOS token outliers.}
A notable cluster of outlier points appears in the low-entropy, high-drift region of each scatter plot (shown in gray). These correspond to denoising steps in which the model has already generated its complete response and is filling the remaining positions with EOS tokens. Because EOS tokens are produced with near-zero entropy yet cause a non-trivial representational shift as the sequence transitions from active generation to padding, they decouple from the entropy--drift trend that governs content tokens. Crucially, these outliers have no practical impact on EntropyCache: since their entropy falls well below any reasonable threshold $\tau$, they are automatically classified as skip steps, and the method correctly reuses the cached KV states. We therefore report all Spearman correlations on non-EOS steps to reflect the regime in which the entropy trigger is operative.

\paragraph{Per-sample Spearman distribution.}
To assess the consistency of the entropy--drift relationship at the individual sample level, Figure~\ref{fig:spearman_distribution} shows the distribution of per-sample Spearman $\rho$ values (non-EOS, log--log) across the 64 samples for each dataset. GSM8K exhibits the tightest distribution (mean $\rho = 0.62$, IQR $\approx 0.59$--$0.68$), while MATH500 (mean $\rho = 0.57$) and MBPP (mean $\rho = 0.59$) show similarly concentrated positive correlations. HumanEval yields a lower and more dispersed distribution (mean $\rho = 0.40$, IQR $\approx 0.33$--$0.50$), indicating that the entropy signal is a weaker predictor of raw KV drift for code generation sequences.

\paragraph{Weak correlation does not imply weak performance.}
Interestingly, the weaker entropy--drift correlation on HumanEval does not translate
into weaker downstream performance. As shown in Table~\ref{tab:main_result_basic},
EntropyCache achieves its highest accuracy improvement over the baseline on HumanEval
(+2.44\%p for LLaDA, +5.49\%p for Dream), despite the comparatively modest $\rho$.
We note that Spearman correlation measures \emph{synchronous} alignment between entropy
and drift at each step: a recomputation triggered 1--2 steps after a drift event would
appear as a mismatch in this metric, yet may still refresh the KV cache before it
affects downstream decoding. Point-wise correlation may therefore underestimate the
practical utility of the entropy signal as a recomputation trigger, and we leave a
deeper mechanistic analysis of this relationship to future work.

\section{Additional validation experiments}
\label{app:additional_validation}

This section reports additional validation experiments: a full component ablation, matched operating-point comparisons, age-conditioned feature drift, batched decoding, and a vocabulary-scaling microbenchmark. All accuracy--throughput experiments use the same unified implementation and decoding settings as the main evaluation unless otherwise specified.

\subsection{Full component ablation}

Table~\ref{tab:component_ablation_full} reports the component ablation on every standard benchmark. The full method has the highest four-task average accuracy among the caching variants on both models. Removing recent-token recomputation particularly affects HumanEval and several Dream tasks, while removing entropy-triggered full refreshes lowers the average on both models. The expected throughput increase of either ablation reflects less recomputation, but comes at the cost of lower average accuracy.

\begin{table*}[!ht]
\centering
\caption{Full component ablation. Each cell reports accuracy (\%) / throughput (tok/s). No cache uses the same confidence-based parallel decoder with full recomputation at every step.}
\label{tab:component_ablation_full}
\resizebox{0.88\textwidth}{!}{%
\begin{tabular}{@{}llcccc@{}}
\toprule
\textbf{Model} & \textbf{Dataset} & \textbf{No cache} & \textbf{Full EntropyCache} & \textbf{Entropy only ($k=0$)} & \textbf{Recent only ($\tau=\infty$)} \\
\midrule
\multirow{5}{*}{LLaDA-8B-Instruct}
& GSM8K     & 78.5 / 8.6  & 78.8 / 38.5 & 76.6 / 42.7 & 74.1 / 47.9 \\
& MATH500   & 42.4 / 9.5  & 42.8 / 39.6 & 39.2 / 43.0 & 40.8 / 52.7 \\
& MBPP      & 42.4 / 16.4 & 38.8 / 64.9 & 39.0 / 71.0 & 40.2 / 78.2 \\
& HumanEval & 48.2 / 15.4 & 48.8 / 38.4 & 42.1 / 43.2 & 44.5 / 44.0 \\
& Average   & 52.9 / 12.5 & \textbf{52.3} / 45.4 & 49.2 / 49.9 & 49.9 / 55.7 \\
\midrule
\multirow{5}{*}{Dream-7B-Instruct}
& GSM8K     & 74.9 / 12.1 & 74.3 / 48.3 & 72.6 / 52.9  & 71.3 / 63.1 \\
& MATH500   & 44.0 / 15.9 & 43.2 / 62.5 & 38.6 / 70.2  & 43.0 / 82.8 \\
& MBPP      & 48.6 / 29.7 & 48.8 / 97.8 & 44.2 / 115.1 & 47.4 / 115.2 \\
& HumanEval & 53.1 / 38.7 & 58.5 / 119.0 & 56.1 / 138.8 & 56.7 / 140.6 \\
& Average   & 55.1 / 24.1 & \textbf{56.2} / 81.9 & 52.9 / 94.3 & 54.6 / 100.4 \\
\bottomrule
\end{tabular}%
}
\end{table*}

\subsection{Matched operating-point comparison}

Default configurations may represent different points on each method's accuracy--efficiency curve. We therefore sweep $\tau$ for EntropyCache, rollout proportion for $\text{d}^2\text{Cache}$, and cosine-similarity threshold for Elastic-Cache on GSM8K (Table~\ref{tab:matched_operating_points}). On LLaDA, the paper EntropyCache setting is faster than every measured $\text{d}^2\text{Cache}$ configuration while attaining comparable measured accuracy, and the EntropyCache frontier dominates the measured Elastic-Cache frontier. At similar recompute ratios of 0.113--0.120, EntropyCache also gives the highest throughput (38.49 vs. 34.5 and 27.1 tok/s). On Dream, the EntropyCache and $\text{d}^2\text{Cache}$ frontiers cross, but EntropyCache remains faster at comparable accuracy and dominates the measured Elastic-Cache points. These sweeps confirm that the advantage is not specific to the paper defaults.

\begin{table*}[!ht]
\centering
\caption{Accuracy--efficiency sweeps on GSM8K under matched prompts, decoding, and hardware. RR denotes recompute ratio; bold rows are the paper configurations.}
\label{tab:matched_operating_points}
\scriptsize
\setlength{\tabcolsep}{3.5pt}
\begin{minipage}[t]{0.49\textwidth}
\centering
\textit{LLaDA-8B-Instruct}\\[2pt]
\begin{tabular}{@{}llrrr@{}}
\toprule
Method & Configuration & Acc. & RR & T-put \\
\midrule
EntropyCache & $\tau=2.5$ & 74.3 & 0.080 & 44.1 \\
EntropyCache & $\tau=2.0$ & 76.1 & 0.088 & 42.0 \\
\textbf{EntropyCache} & $\boldsymbol{\tau=1.5}$ & \textbf{78.77} & \textbf{0.114} & \textbf{38.49} \\
EntropyCache & $\tau=1.0$ & 79.4 & 0.191 & 30.2 \\
EntropyCache & $\tau=0.5$ & 79.8 & 0.489 & 16.1 \\
\midrule
$\text{d}^2\text{Cache}$ & rollout $=0.10$ & 77.0 & 0.093 & 38.1 \\
$\text{d}^2\text{Cache}$ & rollout $=0.15$ & 77.4 & 0.120 & 34.5 \\
\textbf{$\text{d}^2\text{Cache}$} & \textbf{rollout $=0.20$} & \textbf{78.01} & \textbf{0.147} & \textbf{29.96} \\
$\text{d}^2\text{Cache}$ & rollout $=0.25$ & 77.9 & 0.178 & 26.1 \\
$\text{d}^2\text{Cache}$ & rollout $=0.30$ & 78.2 & 0.201 & 24.3 \\
\midrule
Elastic-Cache & similarity $=0.70$ & 74.1 & 0.068 & 30.3 \\
Elastic-Cache & similarity $=0.80$ & 75.3 & 0.090 & 28.6 \\
Elastic-Cache & similarity $=0.85$ & 78.1 & 0.113 & 27.1 \\
\textbf{Elastic-Cache} & \textbf{similarity $=0.90$} & \textbf{79.15} & \textbf{0.140} & \textbf{23.49} \\
Elastic-Cache & similarity $=0.95$ & 79.4 & 0.233 & 20.8 \\
\bottomrule
\end{tabular}
\end{minipage}\hfill
\begin{minipage}[t]{0.49\textwidth}
\centering
\textit{Dream-7B-Instruct}\\[2pt]
\begin{tabular}{@{}llrrr@{}}
\toprule
Method & Configuration & Acc. & RR & T-put \\
\midrule
EntropyCache & $\tau=2.5$ & 71.7 & 0.083 & 56.3 \\
EntropyCache & $\tau=2.0$ & 73.2 & 0.094 & 53.4 \\
\textbf{EntropyCache} & $\boldsymbol{\tau=1.5}$ & \textbf{74.30} & \textbf{0.123} & \textbf{48.32} \\
EntropyCache & $\tau=1.0$ & 74.5 & 0.190 & 39.8 \\
EntropyCache & $\tau=0.5$ & 75.0 & 0.372 & 26.7 \\
\midrule
$\text{d}^2\text{Cache}$ & rollout $=0.10$ & 73.9 & 0.091 & 50.4 \\
$\text{d}^2\text{Cache}$ & rollout $=0.15$ & 74.6 & 0.118 & 44.6 \\
\textbf{$\text{d}^2\text{Cache}$} & \textbf{rollout $=0.20$} & \textbf{75.44} & \textbf{0.145} & \textbf{41.68} \\
$\text{d}^2\text{Cache}$ & rollout $=0.25$ & 75.0 & 0.175 & 38.5 \\
$\text{d}^2\text{Cache}$ & rollout $=0.30$ & 74.6 & 0.198 & 34.2 \\
\midrule
Elastic-Cache & similarity $=0.70$ & 67.8 & 0.064 & 38.1 \\
Elastic-Cache & similarity $=0.80$ & 68.3 & 0.085 & 36.3 \\
Elastic-Cache & similarity $=0.85$ & 69.4 & 0.110 & 34.0 \\
\textbf{Elastic-Cache} & \textbf{similarity $=0.90$} & \textbf{69.60} & \textbf{0.152} & \textbf{32.15} \\
Elastic-Cache & similarity $=0.95$ & 71.3 & 0.235 & 26.6 \\
\bottomrule
\end{tabular}
\end{minipage}
\end{table*}

\subsection{Multi-step feature volatility}

The trajectory examples in Section~\ref{sec:recent_token_volatility} qualitatively motivate recomputation beyond the immediately preceding step. To test this at scale, we aggregate value-vector cosine distance by token age over 64 samples from each standard benchmark (Table~\ref{tab:recency_drift}). Drift falls sharply after unmasking, but remains elevated for several subsequent steps: at age 5 it is still approximately 3--5$\times$ the age-40 level across all tasks. This long-tail behavior supports a multi-step recent-token budget rather than one-step delayed recomputation. The downstream $\tau$--$k_\mathrm{recent}$ grids in Appendix~\ref{app:extended_ablation} provide complementary evidence that moderate recent-token budgets improve the accuracy--throughput trade-off.

\begin{table}[!ht]
\centering
\caption{Mean value-vector cosine distance by token age, measured over 64 samples per dataset with LLaDA-8B-Instruct. Age is the number of denoising steps since unmasking.}
\label{tab:recency_drift}
\resizebox{0.82\textwidth}{!}{%
\begin{tabular}{@{}lrrrrrrr@{}}
\toprule
\textbf{Dataset} & \textbf{Age 1} & \textbf{Age 2} & \textbf{Age 3} & \textbf{Age 5} & \textbf{Age 10} & \textbf{Age 20} & \textbf{Age 40} \\
\midrule
GSM8K     & 0.617 & 0.083 & 0.044 & 0.029 & 0.020 & 0.015 & 0.011 \\
MATH500   & 0.644 & 0.129 & 0.074 & 0.047 & 0.030 & 0.019 & 0.013 \\
MBPP      & 0.659 & 0.137 & 0.080 & 0.052 & 0.023 & 0.016 & 0.012 \\
HumanEval & 0.644 & 0.121 & 0.061 & 0.040 & 0.025 & 0.020 & 0.012 \\
\bottomrule
\end{tabular}%
}
\end{table}

\subsection{Batched decoding}

We evaluate the indexed in-place KV-update implementation under batching on LLaDA-8B-Instruct/GSM8K (Table~\ref{tab:batched_serving}). RTX 6000 Ada GPUs are used because the RTX 3090 setup used in the main experiments does not provide sufficient memory for the larger batches. The nearly flat no-cache and $\text{d}^2\text{Cache}$ throughput curves are consistent with compute saturation: no-cache recomputes all positions, while $\text{d}^2\text{Cache}$ recomputes more positions and incurs attention-rollout overhead. EntropyCache leaves greater batching headroom through lower recomputation and cheaper decisions, increasing throughput by 1.35$\times$ from $B=1$ to $B=8$ while remaining faster at every batch size. During partial passes, selected token positions form the query rows and their newly projected K/V entries are written directly to the same indices in preallocated cache buffers; unchanged entries require no reallocation or full-block copy.

\begin{table}[!ht]
\centering
\caption{Batched GSM8K throughput (tok/s) on LLaDA-8B-Instruct using RTX 6000 Ada GPUs. $B$ is the per-device batch size.}
\label{tab:batched_serving}
\resizebox{0.58\textwidth}{!}{%
\begin{tabular}{@{}lrrrr@{}}
\toprule
\textbf{Method} & $\boldsymbol{B=1}$ & $\boldsymbol{B=2}$ & $\boldsymbol{B=4}$ & $\boldsymbol{B=8}$ \\
\midrule
No cache & 13.95 & 13.55 & 13.02 & 12.66 \\
$\text{d}^2\text{Cache}$ & 61.45 & 62.10 & 60.28 & 59.41 \\
\textbf{EntropyCache} & \textbf{67.30} & \textbf{82.50} & \textbf{90.03} & \textbf{90.99} \\
\bottomrule
\end{tabular}%
}
\end{table}

\subsection{Vocabulary-size scaling}
\label{app:vocab_scaling}

Table~\ref{tab:vocab_scaling} measures the entropy-decision cost across vocabulary sizes. Although the operation is asymptotically $O(NV)$, GPU latency remains 81--122\,$\mu$s for $N\leq 8$ across a 10$\times$ vocabulary range because the small practical regime is dominated by launch overhead. Scaling becomes visible at the conservative $N=32$ upper bound, yet remains below 0.41\,ms at a 320K vocabulary. The measured LLaDA and Dream costs at $N=32$ are only 0.179 and 0.209\,ms, respectively. This is consistent with the end-to-end profile in Figure~\ref{fig:overhead_breakdown}, where entropy decisions account for 71\,ms (0.5\%) of a full generation.

\begin{table}[!ht]
\centering
\caption{Entropy-decision GPU time ($\mu$s) as vocabulary size and the number of newly decoded tokens $N$ vary.}
\label{tab:vocab_scaling}
\resizebox{0.62\textwidth}{!}{%
\begin{tabular}{@{}rlrrr@{}}
\toprule
\textbf{Vocabulary} & \textbf{Representative model} & $\boldsymbol{N=1}$ & $\boldsymbol{N=8}$ & $\boldsymbol{N=32}$ \\
\midrule
32K  & --      & 86 & 82  & 81  \\
126K & LLaDA   & 86 & 81  & 179 \\
152K & Dream   & 92 & 87  & 209 \\
262K & Gemma 2 & 91 & 104 & 340 \\
320K & --      & 92 & 122 & 409 \\
\bottomrule
\end{tabular}%
}
\end{table}

\section{Extended ablation study}
\label{app:extended_ablation}

\subsection{Window size ablation}
\label{app:window_ablation}

Table~\ref{tab:window_ablation} reports accuracy and throughput for EntropyCache ($\tau = 1.5$, $k = 64$) across sliding-window sizes $w \in \{8, 16, 32, 64, 128\}$ on all four standard benchmarks.

For LLaDA-Inst, accuracy remains stable across $w = 8$ to $w = 64$ on most tasks, with $w = 32$ offering the best accuracy--throughput balance: it achieves the highest or near-highest accuracy on GSM8K (78.6\%), MATH500 (42.8\%), and HumanEval (47.6\%), while delivering competitive throughput across all benchmarks. Smaller windows ($w = 8, 16$) preserve accuracy but sacrifice throughput due to the increased number of autoregressive blocks required to cover the generation length.

For Dream-Inst, the sensitivity to window size is more pronounced. Accuracy degrades substantially at $w = 64$ and $w = 128$---for instance, GSM8K drops from 74.3\% ($w = 32$) to 61.9\% ($w = 64$) and 49.7\% ($w = 128$), and MATH500 collapses from 43.2\% to 10.6\% at $w = 128$. This aligns with findings in prior work on semi-autoregressive decoding for dLLMs~\citep{dream, fastdllm}: when the denoising window is large, the model must resolve many tokens simultaneously without the left-to-right bias that smaller windows naturally impose, leading to degraded generation quality. The $w = 32$ configuration strikes the best trade-off between parallel efficiency and the sequential denoising structure that these models benefit from, and we adopt it as the default for all main experiments.

\begin{table}[!ht]
\centering
\caption{Window size ablation for EntropyCache ($\tau = 1.5$, $k = 64$). Acc.: Accuracy (\%). T-put: Throughput (tok/s). Bold indicates the selected configuration ($w = 32$).}
\label{tab:window_ablation}
\resizebox{\textwidth}{!}{%
\begin{tabular}{@{}llcccccccccc@{}}
\toprule
& & \multicolumn{2}{c}{$w = 8$} & \multicolumn{2}{c}{$w = 16$} & \multicolumn{2}{c}{$\mathbf{w = 32}$} & \multicolumn{2}{c}{$w = 64$} & \multicolumn{2}{c}{$w = 128$} \\
\cmidrule(lr){3-4} \cmidrule(lr){5-6} \cmidrule(lr){7-8} \cmidrule(lr){9-10} \cmidrule(lr){11-12}
& \textbf{Benchmark} & Acc. & T-put & Acc. & T-put & Acc. & T-put & Acc. & T-put & Acc. & T-put \\
\midrule
\multicolumn{12}{c}{\textit{LLaDA-Inst}} \\
\midrule
& GSM8K      & 79.08 & 31.66 & 78.92 & 36.68 & \textbf{78.62} & \textbf{38.62} & 77.33 & 35.82 & 68.84 & 28.99 \\
& MATH500    & 41.60 & 31.34 & 40.80 & 37.45 & \textbf{42.80} & \textbf{39.52} & 40.60 & 37.51 & 38.20 & 29.77 \\
& MBPP       & 42.00 & 40.42 & 40.00 & 55.52 & \textbf{38.40} & \textbf{64.53} & 39.20 & 65.71 & 37.80 & 63.93 \\
& HumanEval  & 47.56 & 31.64 & 43.90 & 35.32 & \textbf{47.56} & \textbf{37.86} & 47.56 & 36.25 & 50.00 & 30.72 \\
\midrule
\multicolumn{12}{c}{\textit{Dream-Inst}} \\
\midrule
& GSM8K      & 74.83 & 37.52 & 75.36 & 47.42 & \textbf{74.30} & \textbf{48.06} & 61.87 & 44.67 & 49.66 & 38.16 \\
& MATH500    & 43.80 & 39.10 & 42.20 & 54.85 & \textbf{43.20} & \textbf{62.41} & 33.60 & 55.74 & 10.60 & 41.87 \\
& MBPP       & 53.00 & 52.36 & 56.60 & 79.68 & \textbf{48.80} & \textbf{97.96} & 36.80 & 103.77 & 28.20 & 92.71 \\
& HumanEval  & 54.27 & 52.96 & 60.98 & 86.71 & \textbf{58.54} & \textbf{118.23} & 53.66 & 130.57 & 43.90 & 134.79 \\
\bottomrule
\end{tabular}%
}
\end{table}

\subsection{EntropyCache parameter ablation on extended datasets}
\label{app:param_ablation_extended}

The main text reports the accuracy--throughput grid for GSM8K on both models. Here we present the full per-benchmark ablation over $\tau \in \{0.5, 1.0, 1.5, 2.0, 2.5\}$ and $k \in \{16, 32, 64, 128, 256\}$ on LLaDA-Inst ($w = 32$) in Tables~\ref{tab:ablation_gsm8k}--\ref{tab:ablation_humaneval}.

\paragraph{Entropy threshold $\tau$.}
Across all four benchmarks, accuracy is largely preserved at $\tau \leq 1.5$ and begins to degrade at $\tau \geq 2.0$, while throughput increases monotonically with $\tau$. On GSM8K, accuracy remains within 1\%p of the $\tau = 0.5$ baseline up to $\tau = 1.5$ (e.g., 78.8\% vs.\ 79.8\% at $k = 64$), but drops by ${\sim}3$\%p at $\tau = 2.0$. A similar pattern holds for MATH500 and HumanEval, where $\tau = 1.5$ maintains competitive accuracy while roughly doubling throughput relative to $\tau = 0.5$. The threshold $\tau = 1.5$ thus consistently sits at the knee of the accuracy--throughput curve across tasks.

\paragraph{Recent token budget $k$.}
The effect of $k$ is most visible in throughput: larger $k$ increases the per-step recomputation cost, reducing throughput by 10--15\% when moving from $k = 16$ to $k = 256$ at a fixed $\tau$. On accuracy, the impact is more nuanced. For reasoning-heavy tasks (GSM8K, MATH500), moderate values of $k$ ($32$--$64$) often perform competitively with very small ($k = 16$) or very large ($k = 256$) budgets. For code generation (MBPP, HumanEval), accuracy is relatively insensitive to $k$. Overall, $k = 64$ provides a strong accuracy--throughput balance across all four benchmarks without significantly sacrificing throughput relative to smaller values.

\paragraph{Joint selection.}
Taking both dimensions together, the configuration $\tau = 1.5$, $k = 64$ provides a favorable accuracy--throughput operating point and serves as a shared default across all main experiments.

\begin{table}[!ht]
\centering
\caption{EntropyCache ablation on GSM8K (LLaDA-Inst, $w = 32$). Each cell: Accuracy (\%) / Throughput (tok/s).}
\label{tab:ablation_gsm8k}
\resizebox{0.7\textwidth}{!}{%
\begin{tabular}{@{}lccccc@{}}
\toprule
& $\tau = 0.5$ & $\tau = 1.0$ & $\tau = 1.5$ & $\tau = 2.0$ & $\tau = 2.5$ \\
\midrule
$k = 16$  & 79.76 / 16.12 & 79.15 / 30.87 & 78.47 / 39.44 & 75.44 / 44.34 & 74.30 / 46.26 \\
$k = 32$  & 79.83 / 16.10 & 79.45 / 30.14 & 79.68 / 38.84 & 75.28 / 43.69 & 74.98 / 45.08 \\
$k = 64$  & 79.76 / 16.13 & 79.38 / 30.16 & \textbf{78.77 / 38.56} & 76.12 / 42.04 & 74.30 / 44.07 \\
$k = 128$ & 79.83 / 16.06 & 78.77 / 29.56 & 78.32 / 37.32 & 76.12 / 39.76 & 75.59 / 40.14 \\
$k = 256$ & 79.83 / 16.04 & 78.85 / 29.31 & 78.77 / 35.88 & 75.74 / 37.44 & 74.91 / 37.09 \\
\bottomrule
\end{tabular}%
}
\end{table}

\begin{table}[!ht]
\centering
\caption{EntropyCache ablation on MATH500 (LLaDA-Inst, $w = 32$). Each cell: Accuracy (\%) / Throughput (tok/s).}
\label{tab:ablation_math500}
\resizebox{0.7\textwidth}{!}{%
\begin{tabular}{@{}lccccc@{}}
\toprule
& $\tau = 0.5$ & $\tau = 1.0$ & $\tau = 1.5$ & $\tau = 2.0$ & $\tau = 2.5$ \\
\midrule
$k = 16$  & 41.20 / 17.66 & 41.40 / 30.81 & 41.00 / 42.03 & 39.60 / 47.40 & 39.20 / 50.49 \\
$k = 32$  & 40.80 / 17.65 & 41.40 / 30.44 & 42.20 / 40.40 & 43.00 / 46.13 & 40.60 / 47.90 \\
$k = 64$  & 41.40 / 17.65 & 41.20 / 30.64 & \textbf{42.80 / 39.83} & 43.40 / 45.04 & 39.80 / 46.96 \\
$k = 128$ & 41.20 / 17.48 & 41.20 / 30.18 & 43.00 / 37.78 & 43.20 / 40.82 & 41.00 / 42.44 \\
$k = 256$ & 41.20 / 17.26 & 41.00 / 29.61 & 42.80 / 35.45 & 43.60 / 36.18 & 40.20 / 35.06 \\
\bottomrule
\end{tabular}%
}
\end{table}

\begin{table}[!ht]
\centering
\caption{EntropyCache ablation on MBPP (LLaDA-Inst, $w = 32$). Each cell: Accuracy (\%) / Throughput (tok/s).}
\label{tab:ablation_mbpp}
\resizebox{0.7\textwidth}{!}{%
\begin{tabular}{@{}lccccc@{}}
\toprule
& $\tau = 0.5$ & $\tau = 1.0$ & $\tau = 1.5$ & $\tau = 2.0$ & $\tau = 2.5$ \\
\midrule
$k = 16$  & 42.40 / 33.16 & 41.80 / 53.91 & 38.80 / 65.32 & 40.60 / 70.36 & 38.60 / 72.36 \\
$k = 32$  & 42.60 / 32.79 & 41.80 / 53.75 & 38.80 / 65.19 & 41.20 / 68.88 & 40.60 / 71.97 \\
$k = 64$  & 42.60 / 32.80 & 41.40 / 53.03 & \textbf{38.80 / 63.78} & 40.00 / 68.68 & 40.00 / 69.34 \\
$k = 128$ & 42.60 / 32.20 & 41.60 / 50.95 & 38.80 / 60.85 & 40.60 / 65.73 & 40.80 / 65.63 \\
$k = 256$ & 42.60 / 31.75 & 41.60 / 50.00 & 38.60 / 59.14 & 40.40 / 61.82 & 41.20 / 61.60 \\
\bottomrule
\end{tabular}%
}
\end{table}

\begin{table}[!ht]
\centering
\caption{EntropyCache ablation on HumanEval (LLaDA-Inst, $w = 32$). Each cell: Accuracy (\%) / Throughput (tok/s).}
\label{tab:ablation_humaneval}
\resizebox{0.7\textwidth}{!}{%
\begin{tabular}{@{}lccccc@{}}
\toprule
& $\tau = 0.5$ & $\tau = 1.0$ & $\tau = 1.5$ & $\tau = 2.0$ & $\tau = 2.5$ \\
\midrule
$k = 16$  & 45.73 / 24.55 & 45.73 / 33.87 & 47.56 / 38.97 & 46.34 / 41.73 & 42.07 / 43.91 \\
$k = 32$  & 45.73 / 24.32 & 48.78 / 33.13 & 46.34 / 37.91 & 44.51 / 40.64 & 41.46 / 41.00 \\
$k = 64$  & 45.73 / 24.34 & 48.17 / 33.54 & \textbf{48.78 / 38.10} & 49.39 / 39.45 & 42.07 / 39.46 \\
$k = 128$ & 45.73 / 24.30 & 46.95 / 32.95 & 48.78 / 36.85 & 45.12 / 37.41 & 43.90 / 36.29 \\
$k = 256$ & 45.73 / 24.02 & 47.56 / 31.85 & 49.39 / 35.73 & 47.56 / 33.45 & 46.34 / 29.19 \\
\bottomrule
\end{tabular}%
}
\end{table}

\section{Recompute ratio and throughput analysis}
\label{app:recompute_analysis}

We measure the \emph{recompute ratio}---the fraction of KV states recomputed from scratch per step (baseline $= 1.0$)---to disentangle cache reuse from decision overhead. Table~\ref{tab:recompute_ratio} reports results for the three dynamic methods.

On GSM8K and MATH500, EntropyCache achieves both the lowest recompute ratio and the highest throughput, indicating that the entropy trigger skips more steps without sacrificing cache freshness. On MBPP (LLaDA), EntropyCache recomputes slightly more than $\text{d}^2\text{Cache}$ (0.178 vs.\ 0.164) yet delivers 23\% higher throughput (64.9 vs.\ 52.8 tok/s), demonstrating the wall-clock benefit of its low-cost decision rule over $\text{d}^2\text{Cache}$'s context-length-dependent attention rollout. On HumanEval (LLaDA), EntropyCache has a marginally higher recompute ratio and lower throughput than $\text{d}^2\text{Cache}$, but achieves the highest accuracy among all methods (48.78\% vs. $\text{d}^2\text{Cache}$ \ 41.46\%, Table~\ref{tab:main_result_basic}), suggesting that entropy-triggered recomputation allocates compute to generation-critical steps rather than minimizing recomputation indiscriminately.

\begin{table}[!ht]
\centering
\caption{Recompute ratio (RR) and throughput (T-put, tok/s) for dynamic KV caching methods. Lower RR indicates more cache reuse. Bold: best per benchmark.}
\label{tab:recompute_ratio}
\resizebox{0.6\textwidth}{!}{%
\begin{tabular}{@{}llcccccc@{}}
\toprule
& & \multicolumn{2}{c}{\textbf{Elastic-Cache}} & \multicolumn{2}{c}{\textbf{$\text{d}^2\text{Cache}$}} & \multicolumn{2}{c}{\textbf{Ours}} \\
\cmidrule(lr){3-4} \cmidrule(lr){5-6} \cmidrule(lr){7-8}
& \textbf{Benchmark} & RR & T-put & RR & T-put & RR & T-put \\
\midrule
\multicolumn{8}{c}{\textit{LLaDA-Inst}} \\
\midrule
& GSM8K      & 0.140 & 23.49 & 0.147 & 29.96 & \textbf{0.114} & \textbf{38.49} \\
& MATH500    & 0.148 & 24.43 & 0.146 & 32.18 & \textbf{0.134} & \textbf{39.64} \\
& MBPP       & 0.243 & 32.24 & \textbf{0.164} & 52.77 & 0.178 & \textbf{64.91} \\
& HumanEval  & 0.168 & 27.73 & \textbf{0.160} & 41.36 & 0.172 & 38.41 \\
\cmidrule{2-8}
& Average    & 0.175 & 26.97 & 0.154 & 39.06 & \textbf{0.149} & \textbf{45.36} \\
\midrule
\multicolumn{8}{c}{\textit{Dream-Inst}} \\
\midrule
& GSM8K      & 0.152 & 32.15 & 0.145 & 41.68 & \textbf{0.123} & \textbf{48.32} \\
& MATH500    & 0.148 & 41.90 & 0.147 & 55.80 & \textbf{0.125} & \textbf{62.48} \\
& MBPP       & 0.348 & 54.84 & 0.166 & 87.63 & \textbf{0.145} & \textbf{97.79} \\
& HumanEval  & 0.281 & 72.30 & 0.163 & 110.83 & \textbf{0.160} & \textbf{119.02} \\
\cmidrule{2-8}
& Average    & 0.232 & 50.30 & 0.155 & 73.98 & \textbf{0.138} & \textbf{81.90} \\
\bottomrule
\end{tabular}%
}
\end{table}

\section{Memory usage analysis}
\label{app:memory_analysis}

\begin{figure}[ht!]
    \centering
    \includegraphics[width=0.65\textwidth]{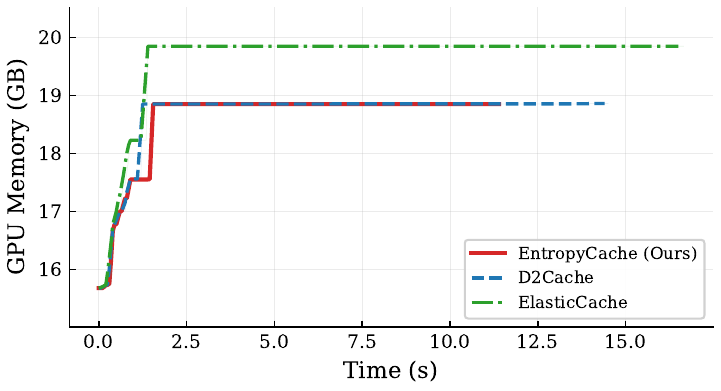}
    \caption{GPU memory usage over time on a single GSM8K sample (LLaDA-Inst, $w = 32$, generation length $= 256$). Elastic-Cache requires ${\sim}$1\,GB more due to extra query and hidden-state caching; $\text{d}^2\text{Cache}$ and EntropyCache are nearly identical.}
    \label{fig:gpu_memory}
\end{figure}

Figure~\ref{fig:gpu_memory} profiles GPU memory usage over time for the three dynamic KV caching methods on a single GSM8K sample (LLaDA-Inst, $w = 32$, generation length $= 256$). Elastic-Cache consumes the most memory (peak 19.85\,GB) because it caches not only key--value pairs but also query projections and hidden states for its per-layer drift comparison. $\text{d}^2\text{Cache}$ and EntropyCache are nearly identical at ${\sim}$18.85\,GB, as neither requires substantial auxiliary storage beyond the KV cache itself. While the absolute differences are modest---model weights and the KV cache dominate the footprint---EntropyCache's auxiliary history costs $O(T)$ memory for $T$ generated positions (one integer per position), independent of prompt length, unlike the context- or depth-dependent buffers of the other methods.

\section{Limitations and broader impacts}
\label{app:limitations}

\paragraph{Limitations.}
As with any approximate inference acceleration method, EntropyCache exhibits accuracy variations across task--model combinations under a single fixed configuration. Our analyses (Section~\ref{sec:ablation}, Appendix~\ref{app:param_ablation_extended}) demonstrate that EntropyCache maintains favorable accuracy--throughput trade-offs across the operating range and outperforms prior caching methods once $\tau$ and $k_\mathrm{recent}$ are appropriately set. We recommend that practitioners calibrate these hyperparameters to their target workload to obtain the best operating point, particularly for tasks such as code generation where token stability dynamics may differ from natural language reasoning.

Additionally, our evaluation covers two dLLM architectures (LLaDA-8B and Dream-7B), the most widely adopted discrete masked diffusion language models in the current research landscape. All prior KV caching works in this space~\citep{d2cache, elasticcache, fastdllm, dkv-cache} evaluate on the same two models. As larger or more diverse dLLMs become available, validating EntropyCache at greater scale would be a natural next step.

\paragraph{Broader impacts.}
By substantially reducing per-request inference cost, EntropyCache can lower the energy and hardware requirements for deploying dLLMs, making diffusion-based generation more accessible in resource-constrained settings. As with any inference acceleration technique, practitioners should verify that approximate caching does not introduce subtle quality regressions in safety-critical applications.

Looking ahead, the entropy-based skipping principle is not limited to the sliding-window decoding setting studied here. Recent block diffusion language models such as Fast-dLLMv2~\citep{fastdllmv2} generate text in coarse blocks that are internally refined through sub-block denoising steps. Since each sub-block undergoes iterative denoising with its own KV dynamics, the entropy trigger proposed in this work could serve as a lightweight recomputation signal within sub-blocks, complementing the block-level KV reuse that these architectures already employ. We leave this extension to future work.

\section{Sample responses}

\subsection{Math}
\label{app:sample_gsm8k}

\begin{figure}[t]
    \centering
    \vfill 
    \includegraphics[width=\textwidth, height=0.9\textheight, keepaspectratio]{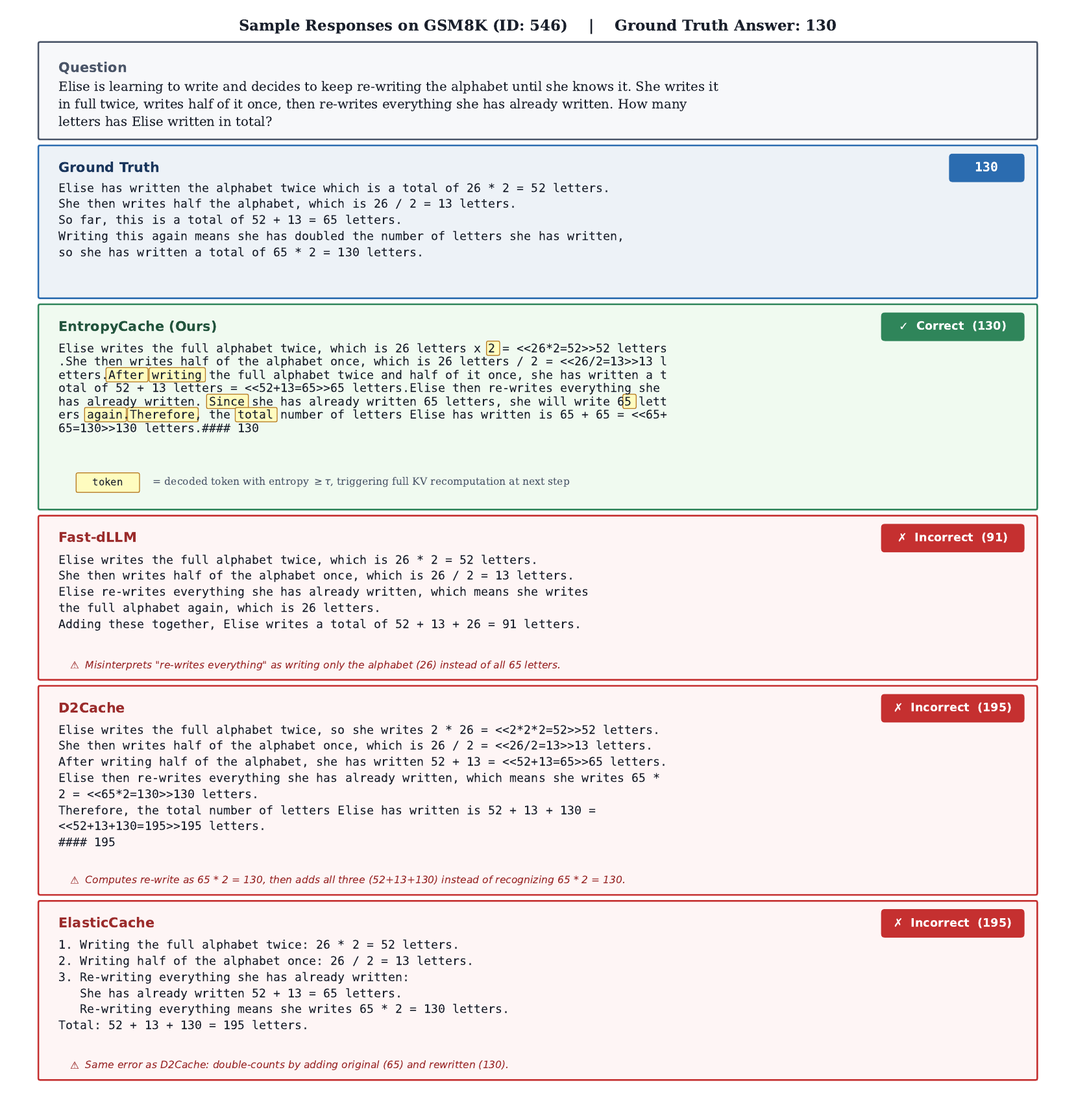}
    \caption{Sample responses on GSM8K (ID: 546, ground truth: 130). Highlighted tokens in the EntropyCache response denote decoded tokens with entropy $\geq \tau$ that triggered full KV recomputation at the next denoising step.}
    \label{fig:sample_gsm8k}
    \vfill 
\end{figure}

Figure~\ref{fig:sample_gsm8k} shows responses from each method on a GSM8K example requiring multi-step cumulative reasoning (ID: 546, answer: 130). Highlighted tokens in the EntropyCache output mark positions where decoded entropy exceeded $\tau$, triggering full KV recomputation. These tend to coincide with transition words and numerically significant tokens. The remaining methods all arrive at incorrect answers.

\subsection{Coding}
\label{app:sample_humaneval}
\begin{figure}[p]
    \centering
    \vfill
    \includegraphics[width=\textwidth, height=0.86\textheight, keepaspectratio]{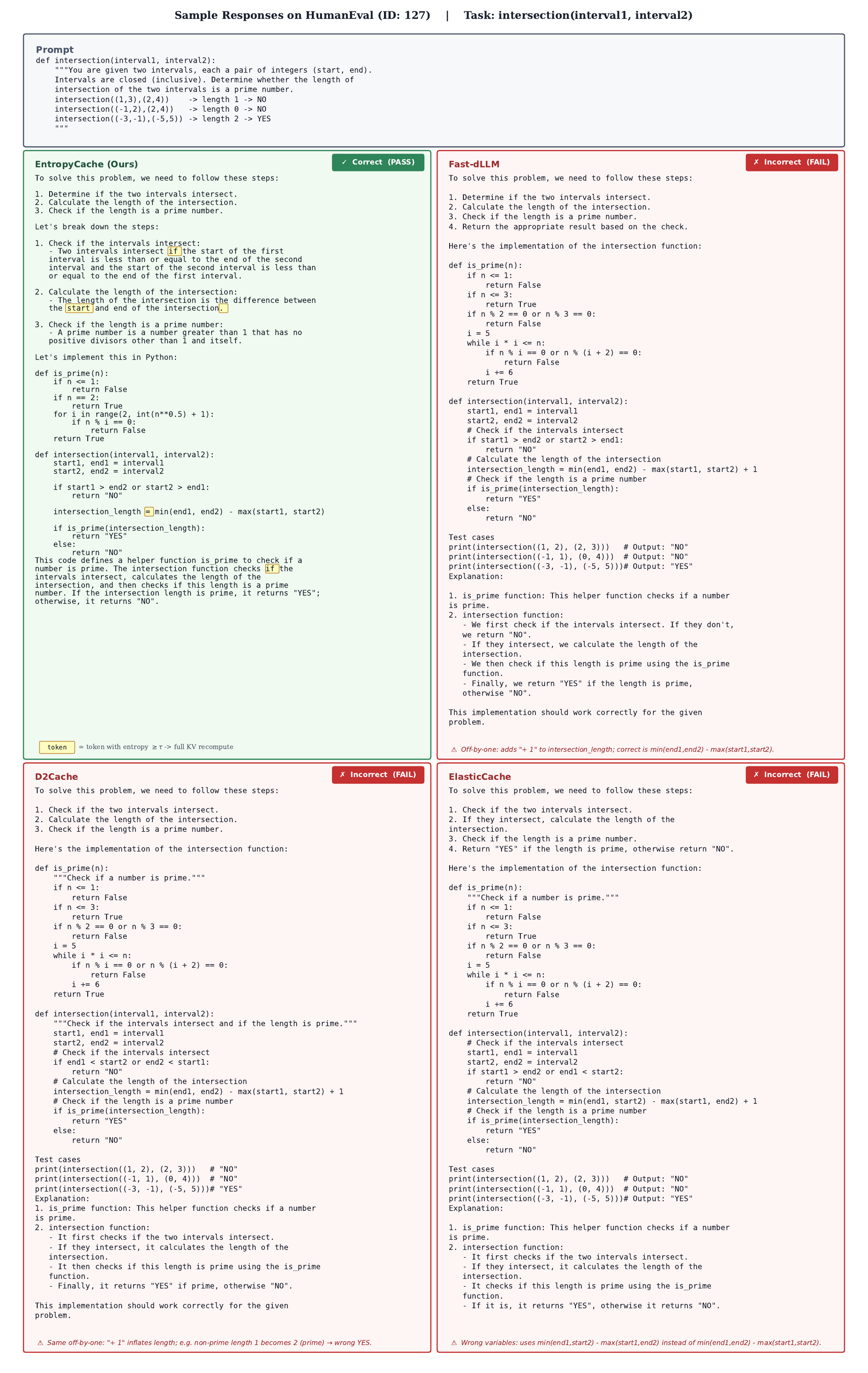}
    \caption{Sample responses on HumanEval (ID: 127). Highlighted tokens in the EntropyCache response denote decoded tokens with entropy $\geq \tau$ that triggered full KV recomputation at the next denoising step.}
    \label{fig:sample_humaneval}
    \vfill
\end{figure}

Figure~\ref{fig:sample_humaneval} shows responses from each method on a HumanEval example requiring interval intersection logic (ID: 127). The task asks whether the length of the intersection of two closed integer intervals is a prime number. Highlighted tokens in the EntropyCache output mark positions where decoded entropy exceeded $\tau$, triggering full KV recomputation. These tend to coincide with logical connectives in the reasoning preamble and the assignment operator in the critical formula. Only EntropyCache produces the correct implementation, while the remaining methods all fail the test cases.

\clearpage

\end{document}